\documentclass{article}

\usepackage[preprint]{neurips_2025}

\usepackage[document]{ragged2e}
\usepackage{amssymb}
\usepackage{amsmath}
\usepackage{amsfonts}
\usepackage{soul,color}
\usepackage{xcolor}
\usepackage{graphicx}
\usepackage{floatrow}
\usepackage{lineno}
\usepackage{float}
\usepackage{hyperref}
\usepackage{comment}
\usepackage{multirow}
\usepackage{booktabs}
\usepackage{geometry}
\usepackage{tabularx}
\usepackage{float}
\usepackage{subcaption}
\usepackage{placeins}

\title{Beyond Attention Heatmaps:\\How to Get Better Explanations for Multiple Instance Learning Models in Histopathology}

\author{%
    \textbf{Mina Jamshidi Idaji}$^{1,2,*,\dagger}$ \quad
    \textbf{Julius Hense}$^{1,2,*}$ \quad
    \textbf{Tom~Neuh{\"a}user}$^{1,2}$ \quad
    \textbf{Augustin~Krause}$^{2}$ \\
    \textbf{Yanqing~Luo}$^{2}$ \quad
    \textbf{Oliver~Eberle}$^{1,2}$  \quad
    \textbf{Thomas~Schnake}$^{3,4,5}$ \quad
    \textbf{Laure~Ciernik}$^{1,2}$  \\
    \textbf{Farnoush~Rezaei~Jafari}$^{1,2}$ \quad
    \textbf{Reza~Vahidimajd}$^{6}$ \quad
    \textbf{Jonas~Dippel}$^{1,2,7}$ \quad
    \textbf{Christoph~Walz}$^{8}$ \\
    \textbf{Frederick~Klauschen}$^{8,9,10}$ \quad
    \textbf{Andreas~Mock}$^{8,9}$ \quad
    \textbf{Klaus-Robert~M{\"u}ller}$^{1,2,11,12}$ \vspace{2mm} \\
  $^{1}$Berlin Institute for the Foundations of Learning and Data, Berlin, Germany \\
  $^{2}$Machine Learning Group, Technische Universität Berlin, Berlin, Germany \\
  $^{3}$ Department of Chemistry, Chemical Physics Theory Group, University of Toronto, Canada \\
  $^{4}$ Vector Institute for Artificial Intelligence, Toronto, Canada \\
  $^{5}$ Acceleration Consortium, University of Toronto, Canada \\
  $^{6}$Department of Computer Science and Engineering, The Chinese University of Hong Kong \\
  $^{7}$Aignostics GmbH, Berlin, Germany \\
  $^{8}$Institute of Pathology, Ludwig Maximilian University, Munich, Germany \\
  $^{9}$German Cancer Research Center, Heidelberg, and German Cancer Consortium, Munich, Germany \\
  $^{10}$Institute of Pathology, Charité Universitätsmedizin, Berlin, Germany  \\
  $^{11}$Department of Artificial Intelligence, Korea University, Seoul, Korea \\
  $^{12}$Max-Planck Institute for Informatics, Saarbrücken, Germany \\
  $^*$Equal contribution \quad
  $^{\dagger}$\texttt{mina.jamshidi.idaji@tu-berlin.de}
}

\begin{document}

\maketitle

\begin{abstract}
\justify
Multiple instance learning (MIL) has enabled substantial progress in computational histopathology, where a large amount of patches from gigapixel whole slide images are aggregated into slide-level predictions.
Heatmaps are widely used to validate MIL models and to discover tissue biomarkers. Yet, the validity of these heatmaps has barely been investigated. In this work, we introduce a general framework for evaluating the quality of MIL heatmaps without requiring additional labels.
We conduct a large-scale benchmark experiment to assess six explanation methods across histopathology task types (classification, regression, survival), MIL model architectures (Attention-, Transformer-, Mamba-based), and patch encoder backbones (UNI2, Virchow2). 
Our results show that explanation quality mostly depends on MIL model architecture and task type, with perturbation (``Single''), layer-wise relevance propagation (LRP), and integrated gradients (IG) consistently outperforming attention-based and gradient-based saliency heatmaps, which often fail to reflect model decision mechanisms. We further demonstrate the advanced capabilities of the best-performing explanation methods: (i) We provide a proof-of-concept that MIL heatmaps of a bulk gene expression prediction model can be correlated with spatial transcriptomics for biological validation, and (ii) showcase the discovery of distinct model strategies for predicting human papillomavirus (HPV) infection from head and neck cancer slides.
Our work highlights the importance of validating MIL heatmaps and establishes that improved explainability can enable more reliable model validation and yield biological insights, making a case for a broader adoption of explainable AI in digital pathology. Our code is provided in a public GitHub repository: \url{https://github.com/bifold-pathomics/xMIL/tree/xmil-journal}.
\end{abstract}
\raggedbottom

\justify
\section{Introduction}
\label{sec:introduction}

Medical machine learning (ML) models frequently operate on heterogeneous and high-dimensional patient input data where only a subset of the available information is relevant to a prediction task. Computational histopathology exemplifies this setting: ML models must identify and aggregate predictive features from gigapixel digital whole slide images (WSIs), which are typically partitioned into thousands of patches, each representing small, potentially informative tissue regions.

This setup is often approached via multiple instance learning (MIL), a popular weakly supervised learning framework in which a label is predicted from a bag of unlabeled instances \citep{elnahhas2025weakly, DIETTERICH199731, ilse2018attentionmil, NIPS1997_82965d4e,pocock2022tiatoolbox, shao2021transmil}.
MIL models can learn to detect diseases \citep{bouzid2024barretts, campanella2019clinical} or classify tumor subtypes \citep{calderaro2023deep, lu2021clam}, aiming to support pathologists in their routine diagnostic workflows. They have further demonstrated remarkable success at tasks that even human pathologists cannot perform reliably due to a lack of commonly known histopathological markers associated with the target. Notable examples include the prediction of clinically relevant molecular biomarkers directly from whole slide images \citep{arslan2024systematic, echle2021biomarkers} such as homologous recombination deficiency (HRD) \citep{schirris22deepsmile, loeffler2024prediction, el2024regression}), micro-satellite instability (MSI) \citep{elnahhas2025weakly, niehues2023generalizable, saillard2023validation, wagner2023transformer} or genetic alterations \citep{campanella2025real, kather2020pan}. Other studies predicted clinical outcomes like survival \citep{liu24advmil, skrede2020outcome}, often from both WSIs and high-dimensional molecular profiles \citep{chen21mcat, eijpe2025dimaf, jaume2024dense}. Recently, MIL has also been used for survival modeling from spatial transcriptomics data \citep{Grouard2025MIL}. 

To expose which features in the input data are relevant, the MIL model prediction can be mapped back into a spatial \textit{heatmap} that highlights important areas in the slide.
Such heatmaps have been used for model validation, i.e., to confirm that the ML model uses tissue features that correspond to known biological patterns and the expectations of human experts \citep{bouzid2024barretts, calderaro2023deep, lu2021ai, jiang2024transformer, song2024analysis}. This is of profound importance in histopathology, where models operating in high-stakes environments are prone to relying on artifacts, staining differences, or demographic confounders instead of biological signal \citep{bender2023towards, hagele2020resolving, howard2021signatures, kauffmann2025unsupervised, koemen2024batcheffects, koemen2025pathorob, lapuschkin2019unmasking, vaidya2024demographic}. Furthermore, heatmaps can enable biomarker discovery, i.e., to uncover previously unknown associations between highlighted features and the prediction target \citep{li2023vision, histo-xai-review, keyl2022patient, keyl2025decoding, mokhtari24interpretable, rodrigues2025imilia}. For example, they can reveal a morphological pattern present in a subgroup of cancer patients predictive of higher survival risk \citep{lee2022derivation}. Another example is the study by \cite{markey2025spatial}, which used heatmaps to identify human-interpretable histopathological markers of TGF$\beta$-CAF, a gene expression signature associated with immunotherapy treatment response.

MIL heatmaps can be computed via various methods (referred to as \textit{explanation~methods}). The most common approach is to extract \textit{attention heatmaps} directly from the attention values of the model, which are readily available in most MIL architectures \citep{boehm2025multimodal, cai2025attrimil, DENG2024103124, ilse2018attentionmil, lu2021clam, shao2021transmil, SHI2024103294}. 
Aside from attention heatmaps, previous works used integrated gradients \citep{lee2022derivation, li2023vision, song2024analysis, vu2025contrastiveig}, layer-wise relevance propagation \citep{hense2024xmil, hense2025pathways, sadafi2023pixelmil}, or perturbation-based methods \citep{early2022shapmil, wagner2023transformer}. Others designed intrinsically interpretable MIL model architectures, where heatmaps can be directly extracted from the prediction scores of the MIL model \citep{javed2022additivemil} (more related work in Section~\ref{supp:sec:related_work}).

Although heatmaps are widely used in MIL, their validity and quality are typically not evaluated. Specifically, it remains unclear how to assess whether MIL heatmaps capture the model strategy and how to systematically compare explanation methods. Yet, the need for heatmap evaluation and explanation method comparison is pressing. First, various works have reported conceptual and practical limitations of attention heatmaps in accurately reflecting model predictions \citep{ali2022xai, jain-wallace-2019-attention, javed2022additivemil, wiegreffe-pinter-2019-attention}. Second, reliance on unvalidated visualizations may lead to confirmation bias, where AI practitioners or medical experts interpret heatmaps in ways that align with their expectations, regardless of whether the explanation faithfully reflects the model output \citep{NAISEH2023102941}. Third, different explanation methods can also yield substantially different heatmaps \citep{hedstrom2023quantus, nauta2023xai_eval, samek2016evaluating}. However, the current literature lacks guidance on when to apply which explanation method.

To address this gap, we present a framework to systematically evaluate heatmaps and compare explanation methods in multiple instance learning, and conduct a large-scale benchmarking study across a wide range of histopathology tasks and model architectures. Specifically, our work makes the following contributions:
\begin{itemize}
\item 
We present \textit{patch flipping} as a practical approach for assessing the faithfulness of MIL heatmaps to the model strategy and provide statistical measures for principled explanation method comparison. Our evaluation framework is model-agnostic and does not require any extra annotations (Section~\ref{sec:methods:evaluating_explanations}).
\item Previous works mostly apply a selected explanation method in the context of a specific application. In contrast, we implement six explanation methods (attention-, perturbation-, and gradient-based) across task types (classification, regression, and survival) and MIL architectures (Attention-, Transformer-, and Mamba-based) (Section~\ref{sec:mil_explanation}). We conduct large-scale experimental comparisons on ten histopathology datasets, revealing that (1) attention heatmaps are unfaithful to the model strategy in almost all cases, (2) the best explanation method depends mostly on task type and architecture, and (3) a group of explanation methods consistently outperformed the others: Single, LRP, and IG (Section~\ref{sec:experiments}).
\item We demonstrate advanced capabilities of the best-performing explanation methods in two practical use cases. First, we establish that heatmaps of a bulk gene expression prediction model can correlate with ground-truth spatial transcriptomics expression, encompassing a novel way of biologically validating MIL models for biomarker prediction (Section~\ref{sec:hest_biological_val}). Second, we combine LRP heatmaps with interpretable tissue features to discover distinct model strategies for predicting human papillomavirus (HPV) infection from head and neck cancer slides (Section~\ref{sec:cell_vit_exploration}).
\end{itemize}

Our work significantly extends our previous conference paper \citep{hense2024xmil}. Regarding our methodology, we generalized the explanation and evaluation methods to be applicable to any task type and added statistical tools for principled explanation method comparison. We further considerably expanded the conceptual and experimental scope by including regression and survival tasks, Mamba-based MIL, six new evaluation datasets, and two practical use cases. Whereas the conference paper introduced xMIL-LRP and formalized the xMIL framework, this manuscript establishes a unified evaluation and selection framework for MIL explanations across architectures and clinical endpoints.

\section{Background: Multiple instance learning (MIL)} \label{sec:background}

In this section, we provide a brief overview of the relevant MIL background.

\subsection{MIL formulations} \label{sec:mil_formulation}

In MIL, a sample is represented by a bag of instances $X = \{ \mathbf{x}_1, \cdots, \mathbf{x}_N \}$ with a bag label $y$, where $\mathbf{x}_n \in \mathbb{R}^{D}$ is the feature vector representing the $n$-th instance. The number of instances per bag $N$ may vary across samples. The associated bag-level label $y$ reflects a target outcome for the entire bag. 
In the original binary formulation \citep{DIETTERICH199731, ilse2018attentionmil, NIPS1997_82965d4e}, it is assumed that binary instance labels $y_{n} \in \{ 0, 1 \}$ exist but are not necessarily known, and the binary bag label is $1$ if and only if at least one instance label is $1$, i.e., $y = \max_{n} \{ {y_{n}} \}$.

Modern MIL applications in histopathology extend this to multi-class classification ($y \in \{1, \dots, C\}$), multi-label settings ($y \in \{0, 1\}^C$), survival prediction (where $y$ encodes time-to-event and censoring), and regression tasks ($y \in \mathbb{R}$).
To reflect the complexities of real-world histopathology tasks, we previously introduced explainable MIL (xMIL, \cite{hense2024xmil}), a more general XAI-based formulation of MIL. Unlike classical MIL formulations, xMIL is not restricted to a specific label structure, but instead assumes that there is some hidden bag-level aggregation function $\mathcal{A}(X) = y$ of arbitrary shape, where $y$ may be categorical, continuous, or time-to-event. In this work, we adopt the general xMIL formulation.

\subsection{MIL pipelines} \label{sec:mil_models}
\begin{figure}[ht!]
    \centering
    \includegraphics[width=\linewidth]{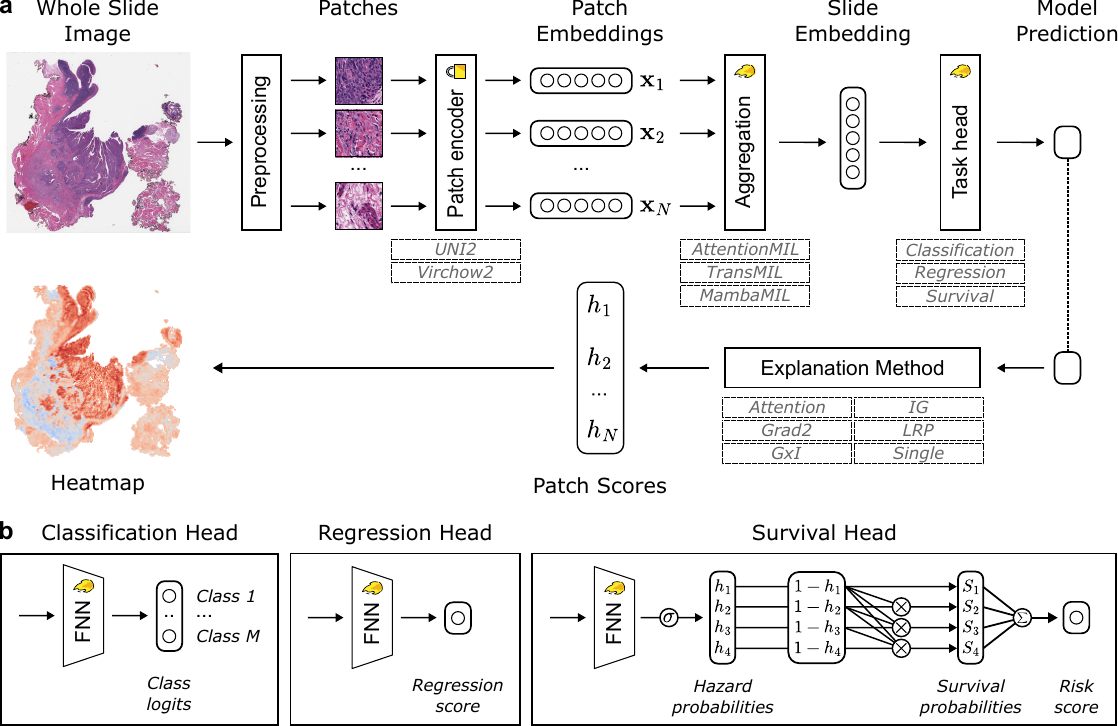}
    \caption{\textbf{Overview of our explanation pipeline for multiple instance learning (MIL)}. \textbf{a}~Block diagram of a general MIL approach to make predictions directly from whole slide images, and to obtain heatmaps to interpret the model predictions. The explanation methods produce patch-wise scores by explaining (only) the task head and the aggregation model, but not the patch encoder. In this work, we assess six explanation methods in the context of two patch encoders, three aggregation model architectures, and three types of prediction heads. \textbf{b} Composition of the task heads. In the survival head, the discrete-time model predicts hazard probabilities $\{h_k\}$ for survival time intervals $k=1, \cdots, K$, and subsequently aggregates them into a patient risk score. See Section~\ref{supp:sec:survival_preliminaries} for details. $\sigma$~=~sigmoid~function; FNN~=~feedforward neural network.}
    \label{fig:mil-model-heads}
\end{figure}

MIL pipelines in histopathology typically consist of three components as illustrated in Figure~\ref{fig:mil-model-heads}-a: a patch encoder extracting embeddings for each patch (=instance) of a whole slide image (=bag), an aggregation model fusing the patch embeddings into a slide embedding, and a task-specific prediction head inferring the final slide-level model prediction. Each model is then trained by optimizing a task-specific loss function. In the following, we describe these components in detail.

\subsubsection{Patch encoder}

The patch encoder module can range from predefined, hand-crafted feature extractors to pretrained deep neural network backbones, including domain-specific foundation models. In histopathology, foundation models have become powerful feature extractors suitable for a wide range of tasks \citep{alber2026atlas2, chen2024uni, dippel2024rudolfv, vorontsov2023virchow, wang2022ctranspath}. The weights of the backbone are often frozen, allowing for a more efficient training of the aggregation block. Optionally, the backbone can be finetuned. In this work, we considered the recent UNI2\footnote{\url{https://huggingface.co/MahmoodLab/UNI2-h}} \citep{chen2024uni} and Virchow2 \citep{zimmermann2024virchow2} foundation models as feature extractors, which have achieved competitive results in previous performance and robustness benchmarks \citep{campanella2025benchmark, filiot2025distilling, jaume2024hest, kaiko2024eva, koemen2025pathorob, mollers2026sampling, neidlinger2025benchmarking}.

\subsubsection{Aggregation}

For the aggregation model (or MIL model), we include three types of widely used architectures, which we consider prototypical MIL models: Attention MIL, Transformer-based MIL, and Mamba-based MIL.

In Attention MIL (AttnMIL) \citep{ilse2018attentionmil}, a popular MIL model in histopathology \citep{el2024regression, Graziani2022reg, lu2021clam, shao2025do}, a bag embedding is computed as an attention-weighted average of instance-level embeddings.
In Transformer-based MIL (TransMIL\footnote{The original Transformer-based MIL model proposed in \citep{shao2021transmil} is also referred to as TransMIL. Here, for brevity, we refer to general Transformer-based MIL as TransMIL, although it may differ in part from the original architecture proposed by \cite{shao2021transmil}.}), instances are treated as tokens and passed through self-attention layers of Transformer blocks, modeling arbitrary pairwise interactions between the instances \citep{kernelmil2021, shao2021transmil}. Several studies in digital histopathology have adopted Transformer-based models for various prediction tasks \citep{chen21mcat, fourkioti2024camil, shao2025do, wagner2023transformer}.
As an alternative to Transformer-based MIL, \cite{yang2024mambamil} proposed MambaMIL, allowing linear scaling in sequence length without resorting to approximations of self-attention. Instance-level embeddings are first processed by a Mamba layer \citep{gu2024mamba}, which models interactions between instances, and then aggregated by attention pooling to obtain a bag-level representation. All of these aggregation models can compute non-linear feature interactions between the instance information, which makes the derivation of meaningful explanations non-trivial.

Numerous variations of the original AttnMIL, TransMIL, and MambaMIL architectures have been proposed (see Section~\ref{supp:sec:related_work}). In this work, we focus on explaining prototypical architectures, expecting that our insights will transfer to such variations.

\subsubsection{Task head}

MIL has been used in almost all types of machine learning tasks on WSI images, including classification, regression \citep{el2024regression, Graziani2022reg}, and survival prediction \citep{chen21mcat, liu24advmil}. Each task requires a task-specific prediction head (overview in Figure~\ref{fig:mil-model-heads}-b), which also demands task-specific adaptations in the explanation methods (see below).

For classification, a shallow feedforward neural network (FNN) --- often just a linear layer --- is used to map the bag representation to the class logits. In regression, the FNN outputs a single node for predicting the continuous target. 
In a survival prediction task, each sample is accompanied by a continuous target defining the event time (overall survival time, disease-free survival time, relapse time, etc.) as well as a binary \textit{censorship} variable, which defines whether the event is observed for this sample. We follow a discrete-time survival modeling where the time span is divided into $K$ time intervals, since it enables effective model training even with small batch sizes \citep{chen21mcat, zadeh2020bias}. Given a bag of instances $X$ associated with the discretized event time $T$, the discrete-time survival probabilities are modeled as $S_k = f_{survival}(k \mid X) = P(T>k  \mid X) = \prod_{u=1}^{k} (1 - h_u)$, where $h_k = f_{hazard}(k \mid X)=P(T=k \mid T\geq k, X)$ is the hazard function and is modeled via a MIL model with a linear head with $K$ outputs connected to sigmoid functions. The final output of a MIL survival model is a risk score $r=-\sum_{k=1}^K f_{survival}(k \mid X)$. In Section \ref{supp:sec:survival_preliminaries}, we provide a more detailed description of the above-mentioned discrete-time survival modeling, comparing it to the Cox proportional hazards model \citep{cox1972regression, katzman2018deepsurv} and details of the log-likelihood cost function for training such a survival model \citep{zadeh2020bias}.

\section{Methods} \label{sec:methods}
\subsection{Overview}
\begin{figure}[t!]
    \centering
    \includegraphics[width=0.5\linewidth]{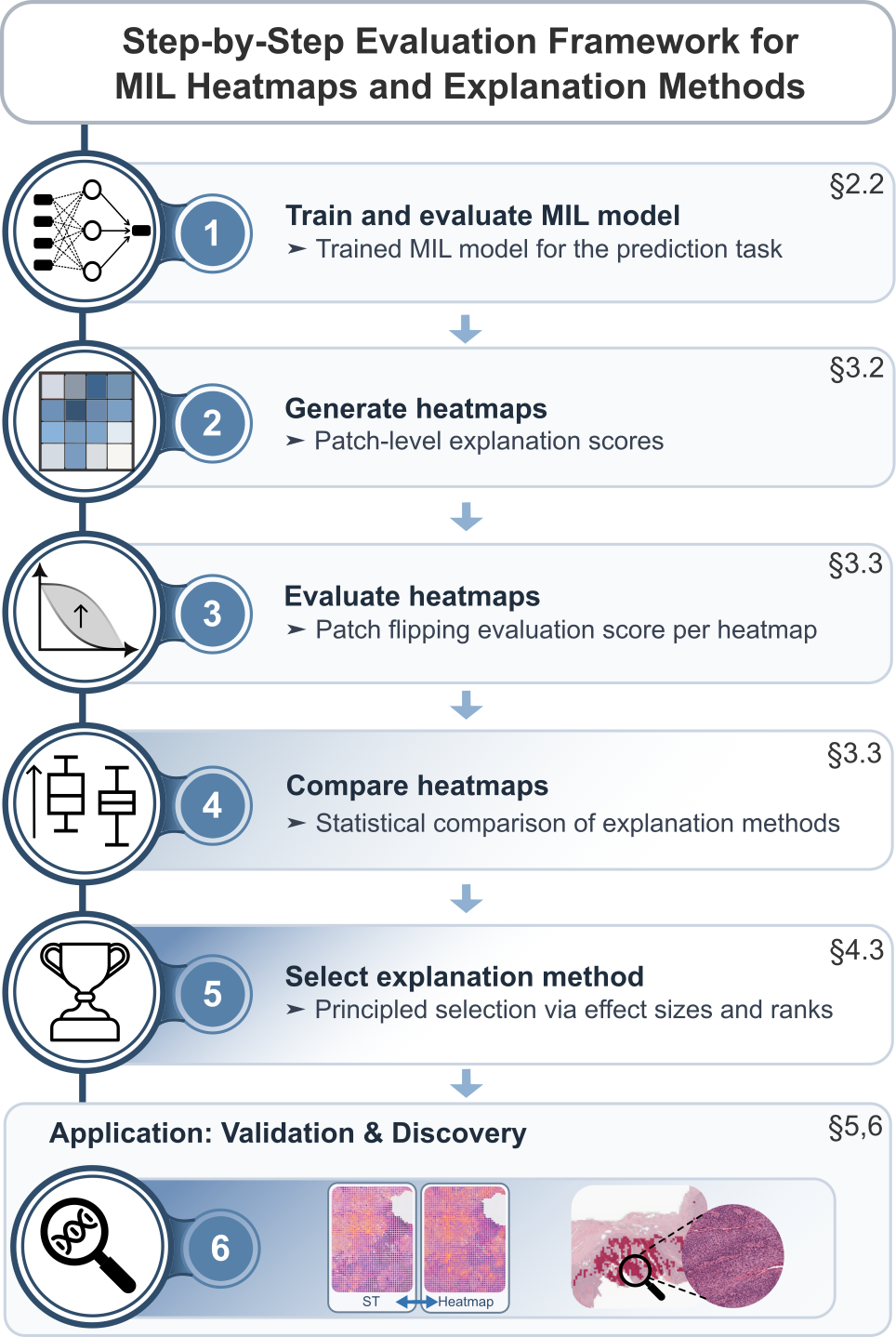}
    \caption{\textbf{Overview of the evaluation framework for comparing heatmaps and explanation methods in multiple instance learning (MIL)}.}
    \label{fig:framework_workflow}
\end{figure}
Before discussing the methodological details, we provide a high-level overview of the evaluation framework proposed in this work.
One of the central objectives of this study is to show how to identify the most reliable method for explaining the predictions of a MIL model. Figure~\ref{fig:framework_workflow} outlines our proposed evaluation pipeline: (1) The process begins with training and evaluating a MIL model for a specific prediction task—classification, regression, or survival (Section \ref{sec:mil_models}).\footnote{We emphasize that in this work, we do not focus on \textit{how} to train good MIL models, but on the explanation of these models. For this, we only require MIL models that achieve a reasonable predictive performance.} (2) For each whole-slide image (WSI), slide-level heatmaps are generated using multiple explanation methods (Section \ref{sec:mil_explanation}). Each heatmap is flattened into an explanation score vector over the patches and (3) evaluated using a patch-flipping perturbation test (Section \ref{sec:methods:evaluating_explanations}): patches are progressively removed based on ascending and descending orderings of their relevance scores, the model predictions are recomputed for the modified bags of patches, and the area between the resulting perturbation curves is calculated as the evaluation score (higher values indicate more \textit{faithful} heatmaps). (4) This score is computed per slide and heatmap, enabling cohort-level statistical comparison of heatmaps via effect size and rank-based analyses (Section \ref{sec:methods:evaluating_explanations}). (5) Based on these comparisons, the best-suited method is selected, and its heatmaps can then be used for qualitative or quantitative downstream analyses, e.g., model validation and generating insights into the model strategy and the underlying biology.

\subsection{Explaining multiple instance learning} \label{sec:mil_explanation}

Explaining a MIL model involves assigning an explanation score to each instance within a bag. In histopathology, where instances correspond to patches extracted from WSIs, these explanation scores can be spatially arranged to form a heatmap overlaid on the original slide (see Figure~\ref{fig:mil-model-heads}-a). Note that only explaining the task head and the aggregation model is sufficient to obtain \textit{instance-level} explanation scores; the patch encoder can be ignored (unless more fine-grained, pixel-level heatmaps are desired). Formally, for an aggregation model $f$ and a bag $X=\{\mathbf{x}_1, \dots, \mathbf{x}_N\} \subset \mathbb{R}^D$, a heatmap is an ordered set $H=\{\epsilon_1, \dots, \epsilon_N\}\subset \mathbb{R}$ which is linked to the bag. That is, there is a heatmapping function $\mathcal{H}_f: X \to H$, such that $\mathcal{H}_f(\mathbf{x}_n | X) = \epsilon_n$. Note that the explanation score (or heatmap score) $\epsilon_n$ depends on the whole bag $X$, i.e., it is context-sensitive.

The xMIL framework from our previous work \citep{hense2024xmil} formally integrates explanation scores into MIL and formulates desiderata for explanation heatmaps. Specifically, heatmaps should be \textit{context-sensitive}, meaning patch relevance depends on both local features and their relations to other patches. They must distinguish \textit{positive from negative evidence}, showing whether regions support or contradict the prediction. Finally, they should preserve \textit{ordering}, with higher scores reflecting stronger contributions to the decision.

\subsubsection{Explanation methods} \label{sec:methods:xA_methods}
In this section, we provide a brief overview of the six explanation methods we employed in this work. These methods cover frequent practices for explaining MIL models in previous works on histopathology applications as well as in XAI research in general. We provide more details in Section~\ref{supp:sec:exp_methods}.

\textit{Attention heatmaps} \textbf{(Attn)} represent the area a MIL model focuses on to make a prediction. The attention heatmap scores are always positive and consist of either the raw attention weights (as in AttnMIL and MambaMIL) or an aggregated version \citep{abnar-zuidema-2020-quantifying} if multiple attention blocks are used (as in Transformer-based MIL).

\textit{Gradient-based} methods are another family of XAI methods and utilize gradient information to derive feature relevance scores.  We use Integrated Gradients \textbf{(IG)} \citep{sundararajan2017axiomatic} as a more sophisticated method, as well as Gradient $\times$ Input \textbf{(G$\times$I)} \citep{baehrens10a,montavon2018methods, DBLP:journals/corr/ShrikumarGSK16} and saliency maps of squared gradients \textbf{(Grad2)} as more naive baselines.

\textit{Perturbation-based} methods estimate instance importance by systematically modifying the input bag and measuring the effect on the model output. We adopt the single-instance approach (referred to as \textbf{Single}), where the model prediction for a bag consisting of a single instance is used as an explanation score for this instance \citep{early2022shapmil}.

\textit{Layer-wise relevance propagation} (\textbf{LRP},  \cite{bach-plos15, montavon2019layer, samek2021-xai}) backpropagates the model’s output relevance to individual patches by redistributing prediction evidence through the network based on relevance conservation rules (see Section \ref{supp:sec:lrp_rules} for details). \cite{hense2024xmil} have adapted LRP to MIL models.

\subsubsection{Task-specific explanations} \label{sec:methods:task_specific_xai}

The task type (classification, regression, or survival) and the corresponding structure of the MIL model task head (see Section~\ref{sec:mil_models}) influence \textit{how} explanations can be computed and \textit{what} information the resulting heatmaps provide.

\paragraph{Classification}
Classification models predict a score for each of the potential classes. This enables computing a class-specific explanation, measuring which areas in the slide are evidence for or against a particular class.
Attention heatmaps highlight focus regions of the MIL model without distinguishing which class these regions are associated with. However, all other explanation methods produce class-specific heatmaps. For gradient-based methods and LRP, the backpropagation starts from one of the model logits. For Single, changes in the softmax score of the selected class are measured.

\paragraph{Regression}
The output of regression MIL models is a real-valued score estimating a continuous target. For a heatmap to be meaningful in this context, \cite{letzgus2022} argued that a user-defined \textit{reference value} is required: the heatmap then indicates which areas in the slide are evidence for a prediction score higher or lower than the reference value.
The reference value can be derived from the training data (e.g., the median score) or chosen from expert knowledge. For instance, the homologous recombination deficiency (HRD) biomarker has a clinically validated cutoff of 42 \citep{el2024regression, loeffler2024prediction}.
For computing heatmaps, we adopt the \textit{retraining} strategy introduced by \cite{letzgus2022}, i.e., (re-)training the MIL regression model to predict the difference between the target and the reference value and explaining this difference prediction. Subsequently, we use the predicted difference score as the starting point for backpropagation in gradient-based methods and LRP, and for observing changes in perturbation-based methods. This approach also affects attention heatmaps: after retraining, they highlight the focus areas of the MIL model to decide whether the prediction score is higher or lower than the reference value.

\paragraph{Survival}
For discrete-time survival models, we aim to explain the risk score (see Section~\ref{sec:mil_models} and Equation~\ref{supp:eq:discrete_survival_model_risk_score} of the supplementary material). In this context, heatmaps outline regions providing evidence for high survival risk or low survival risk; specifically, higher or lower survival risk than the model-inherent reference hazard at $h_{k}=0.5$ for all the time intervals; i.e., 50\% survival chance per survival time interval (see \ref{supp:sec:survival_modeling_details} for details).
We use the risk score as the starting point for backpropagation in gradient-based methods, and for observing changes in perturbation-based methods. For LRP, we developed a composite explanation approach. We first apply Gradient~$\times$~Input to the risk score with respect to the model logits before the sigmoid ($l_k \cdot \partial r / l_k$), which yields a relevance vector of size $K$ corresponding to the hazard logits. This relevance vector is then propagated back to the input patches using the LRP rules (see Section~\ref{sec:methods:xA_methods}).
Notice that for attention heatmaps, it is conceptually unclear how the focus areas correspond to the predicted survival risk.

\subsection{Evaluating MIL explanations}\label{sec:methods:evaluating_explanations}

To assess how accurately a heatmap reflects the model prediction, we adapted the concept of \textit{faithfulness} from XAI research to the MIL setting \citep{blucher2024flipping, hense2024xmil, samek2016evaluating}. The original idea of a faithfulness experiment is to iteratively occlude features in the input based on their explanation score ranking and measure the resulting change in the model prediction. Crucially, this \textit{occlusion} should not cause the input to go out of distribution, which makes the choice of inpainting method essential  (i.e., how to replace an occluded feature). We argue that this challenge does not arise in MIL: since a MIL model naturally handles varying bag sizes, patches can simply be removed from the bag without creating an out-of-distribution scenario. Therefore, we progressively remove patches from a bag and record the resulting model predictions to compute an evaluation metric that quantifies the heatmap quality (referred to as \textit{patch flipping}). Notably, this evaluation framework does not require any ground truth knowledge and can always be employed to assess how well a heatmap reflects the model output in any specific application. 

\subsubsection{Patch flipping}

\begin{figure}[t!]
    \centering
    \includegraphics[width=\linewidth]{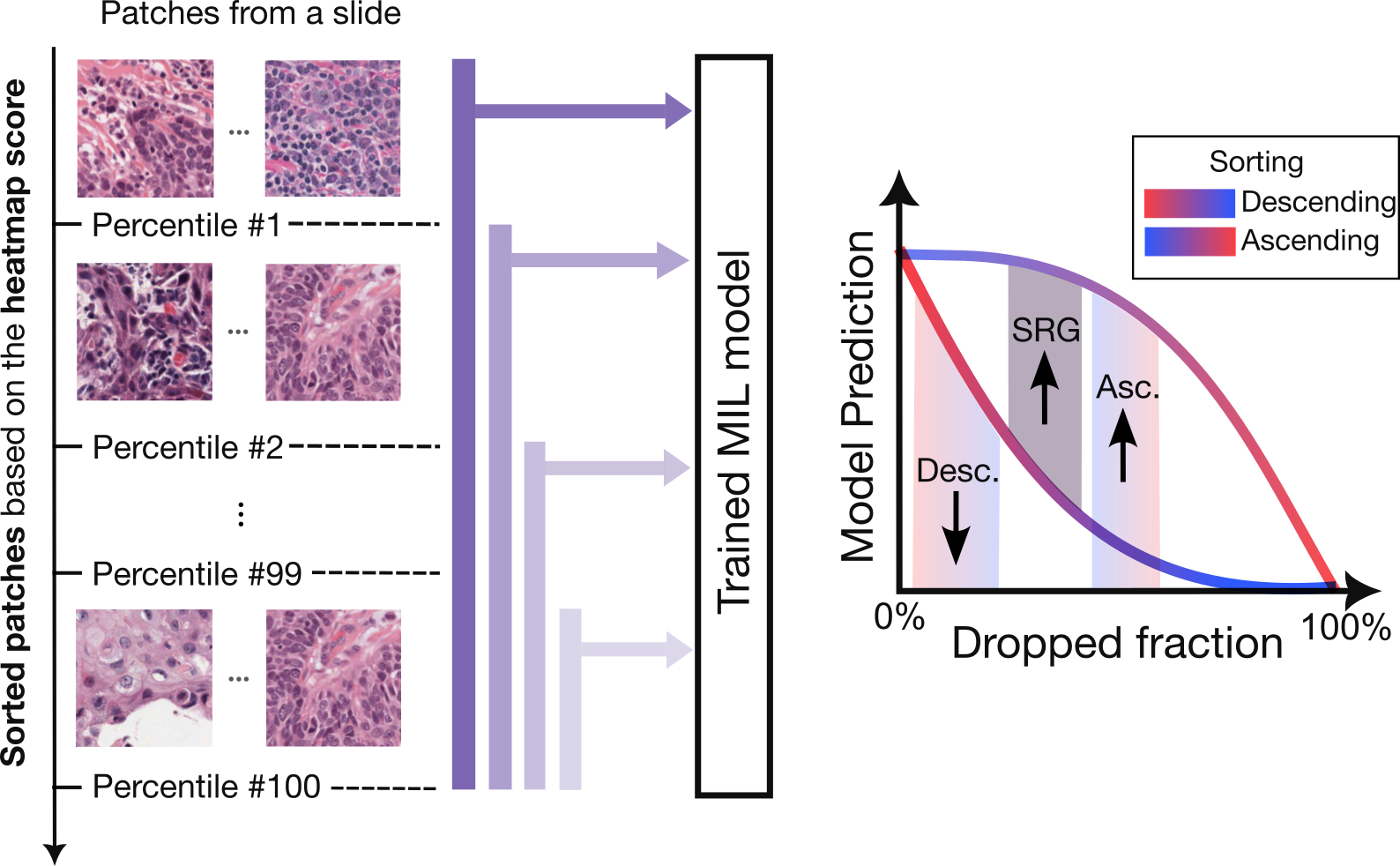}
    \caption{\textbf{The patch flipping procedure}. Patches are sorted based on their heatmap scores either in ascending or descending order. They are then grouped into 100 mini-bags. The mini-bags are removed progressively from the main bag of patches, and the remaining patches are fed into the model. Two perturbation curves (model prediction vs.\ dropped fraction) are formed for the two orderings. A larger area under the curve (AUC) of the ascending (Asc.) curve is desired, whereas a smaller AUC of the descending (Desc.) curve is better. Therefore, the area between the two curves (called Symmetric Relevance Gain--SRG) quantifies the faithfulness of the heatmap. A larger SRG indicates a more faithful heatmap.}
    \label{fig:patch_flipping_visualization}
\end{figure}

Figure \ref{fig:patch_flipping_visualization} visualizes the patch flipping procedure. Given a bag $X=\{\mathbf{x}_1, \cdots, \mathbf{x}_N\}$ and a heatmap function $\mathcal{H}$ that produces explanation scores for the instances in $X$, i.e., $\mathcal{H}(\mathbf{x}_n)=\epsilon_{n}$, we first rank the patches using a ranking function $\pi$ and then partition them into 100 ordered sets $E^{\pi}=(E_1^{\pi}, \cdots, E_{100}^{\pi})$. $E_i^{\pi}$ contains all patches whose ranked scores fall between the $(100-i)$-th and $(100-i+1)$-th percentiles of the ranked explanation scores $\{\pi(\epsilon_n)\}$. For example, $E_1^{\pi}$ is the set of top 1\% of patches according to the ranking $\pi$.

Following the region perturbation strategy introduced by \cite{samek2016evaluating} and \citep{blucher2024flipping}, we progressively exclude the patches based on the chosen ordering $\pi$. Starting from $E_1^{\pi}$ and moving sequentially to $E_{100}^{\pi}$, we remove each partition and monitor the model's output. Formally, the perturbation procedure is defined as:
\begin{equation}\label{eq:perturbed_inputs}
\begin{gathered}
X_0^{\pi} = X \\
X_m^{\pi} = \bigcup_{i=m+1}^{100}E_i^{\pi}, \quad m=1, \cdots, 100 \\
\end{gathered}
\end{equation}
For $m=100$, all patches are removed and therefore, $X^{(100)}$  is an empty set, for which we pass an array of zeros to the model.

To comprehensively assess the effect of explanation score ordering, we perform this partitioning for two orderings: once with the explanation scores sorted in descending order (most relevant to least relevant) and once in ascending order (least relevant to most relevant).

Using the perturbed bags, defined in Equation \ref{eq:perturbed_inputs}, we can construct a curve $f(X_i^{\pi})$ vs.\ $i$, for each ordering $\pi$. The area under this perturbation curve (AUPC) is then computed as:
\begin{equation} \label{eq:aupc}
    \text{AUPC}(X, \mathcal{H}, \pi) = \frac{1}{100} \sum_{m=0}^{100} f(X_m^{\pi})
\end{equation}
where $f$ is the model's function (aggregation model + task head).

For a faithful heatmap $\mathcal{H}$, we expect a rapid change in model output when the most relevant patches are removed first; therefore, a low $\text{AUPC}(X, \mathcal{H}, \pi=\text{descending})$ is desired. Conversely, when the least relevant patches are removed first, the model output should change more slowly, implying that a higher $\text{AUPC}(X, \mathcal{H}, \pi=\text{ascending})$ is preferred. Consequently, \cite{blucher2024flipping} define the Symmetric Relevance Gain (SRG) as a metric for evaluating heatmap quality, which captures the gap between these two perturbation curves:
\begin{equation}\label{eq:SRG}
    \text{SRG}(X, \mathcal{H}) = \text{AUPC}(\pi=\text{ascending}) - \text{AUPC}(\pi=\text{descending})
\end{equation}
A larger value for SRG indicates a more faithful heatmap. In the above equation, $\text{AUPC}(\pi)$ denotes the shorthand notation for $\text{AUPC}(X, \mathcal{H}, \pi)$ for brevity.

\subsubsection{Statistical comparison of explanation methods}
\label{sec:methods:statistical_comparison}

The patch flipping analysis results in one SRG score for each heatmap. For an aggregated faithfulness comparison between explanation methods across a whole test cohort of WSIs, we propose two complementary analyses, as described below. In both cases, we adopt a within-subject experimental design, where each slide provides SRG scores for multiple explanation methods (repeated measurements).

\paragraph{Pairwise comparisons}
We perform a pairwise comparison between explanation methods, allowing us to quantify performance gaps. This approach enables us to group methods with similar levels of faithfulness.
We test the pairwise differences among the explanation methods within each experimental configuration using Wilcoxon's signed rank test. We report the $p$-values (FDR corrected) and test's effect size \citep{RAMACHANDRAN2021491}. With this approach, we are able to aggregate the effect sizes across different experimental settings.

The sign of the effect size indicates whether the first method yields higher SRG scores than the second (positive sign) or lower (negative sign). As an inferential statistical tool, we adopt conventional thresholds used for Cohen’s d ($|r| < 0.2$: negligible; $0.2–0.5$: weak to moderate; $> 0.5$: strong, with $r$ referring to the effect size).

To verify that our conclusions are not driven by a specific effect size definition, we additionally repeat all aggregated pairwise analyses using a robust standardized median based on the median absolute deviation (median/MAD) (see Section~\ref{supp:sec:stat_details}).

\paragraph{Rank-based comparisons}
To accentuate fine-grained differences between explanation methods for method selection, we additionally adopt a within-subject ranking strategy. For each slide, we assign a rank to each explanation method according to its SRG score (highest SRG = rank 1, lowest SRG = last rank). In our experiments, the maximum rank can be seven (six methods + random baseline). We then average these rank scores across all slides in the test set, resulting in a \textit{Mean Rank Score} (MRS) per explanation method. A lower MRS for an explanation method indicates that it was ranked better on average for all slides, independent of the effective differences across explanation methods. The MRS allows us to identify subtle but consistent quality differences between the heatmaps, even if they are small in absolute effect size. Compared to pairwise statistical tests with continuous effect sizes, the rank-based approach discretizes the differences among the explanation methods and, therefore, provides a more straightforward selection approach. 

\section{Large-scale evaluation of MIL heatmaps} \label{sec:experiments}

\begin{figure}[t!]
    \centering
    \includegraphics[width=\linewidth]{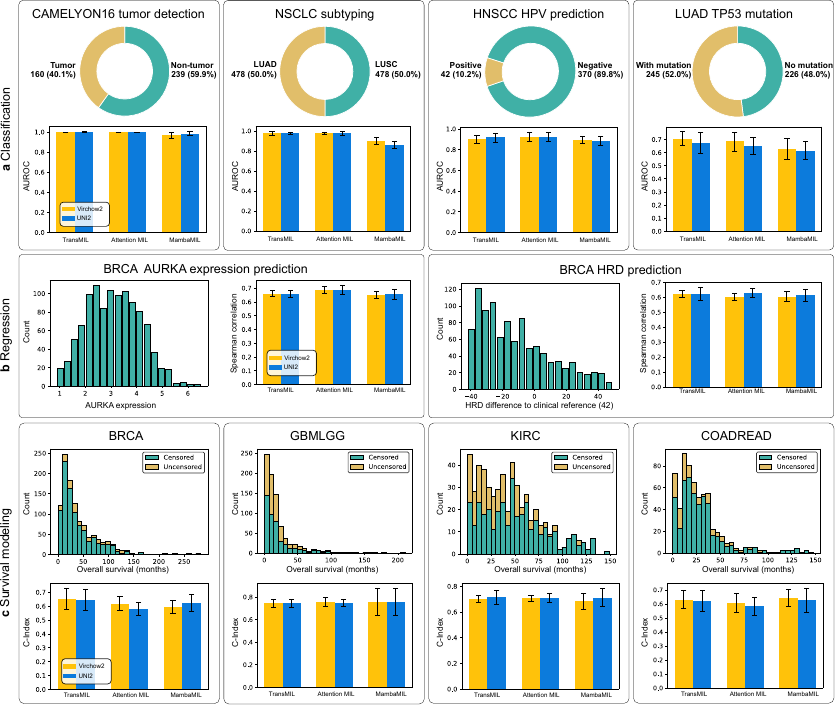}
    \caption{\textbf{Datasets and model performances.} The bar plots in each panel show the mean test performance over the cross-validation folds with the error bars depicting the standard deviations (see Tables~\ref{supp:tab:model_performances_virchow} and ~\ref{supp:tab:model_performances_uni} for numerical model performance results). For the classification tasks in panel (a), the ratio of the cases in each class is depicted. In panels (b) and (c), the endpoint target's histograms are illustrated.}
    \label{fig:model_dataset_figure}
\end{figure}
\subsection{Experimental design}

We designed a repeated-measures experimental study to evaluate the faithfulness of the heatmaps of six explanation methods (see Section~\ref{sec:methods:xA_methods}) across a wide range of histopathology applications and MIL pipelines. Specifically, we considered three predictive task types: classification, regression, and survival analysis. For each task type, multiple datasets were used, covering various practically relevant use cases and levels of difficulty (overview in Figure~\ref{fig:model_dataset_figure}, details in Section~\ref{supp:sec:experimental_setup}).

We extracted patches at 20x magnification from WSI tissue regions and derived patch-level feature vectors from the (frozen) UNI2 and Virchow2 foundation models. We trained AttnMIL, TransMIL, and MambaMIL models (see Section~\ref{sec:mil_models}) using hyper-parameter optimization and a 5-fold cross-validation scheme. We applied the best-performing model checkpoints to held-out test sets, and computed heatmaps for all explanation methods on all test set slides. For each heatmap, we then conducted a patch flipping experiment (see Section~\ref{sec:methods:evaluating_explanations}) and used the resulting SRG scores to compare the faithfulness of the explanation methods as described in Section~\ref{sec:methods:statistical_comparison}. Experimental details are provided in Section~\ref{supp:sec:experimental_setup}.

\subsection{Model performances}

In this work, our objective is not to compare model performance with prior studies but to evaluate the explanation methods themselves. Consequently, small differences in model performances are not central to our analysis, as a sufficiently well-performing model is adequate for our purposes.

Since our evaluation tasks covered different levels of difficulty, the mean model performances varied strongly between tasks (Figure~\ref{fig:model_dataset_figure}, Tables~\ref{supp:tab:model_performances_virchow},~\ref{supp:tab:model_performances_uni}). In CAMELYON tumor detection, for example, almost perfect AUCs were reached (0.97-1.00), whereas LUAD TP53 mutation prediction AUCs only ranged from 0.62-0.71.
The performance differences between backbones and MIL models were mostly small and often negligible. TransMIL sometimes yielded considerably better results (e.g., in LUAD TP53 classification, BRCA survival prediction), showcasing its high performance potential. At the same time, MambaMIL and TransMIL sometimes reached considerably lower performance than AttnMIL, likely due to higher sensitivity to (suboptimal) hyperparameter choices.

\subsection{Comparison of MIL heatmaps in example use cases} \label{sec:results:faithfulness_examples}

\begin{figure}[t!]
    \centering
    \includegraphics[width=\linewidth]{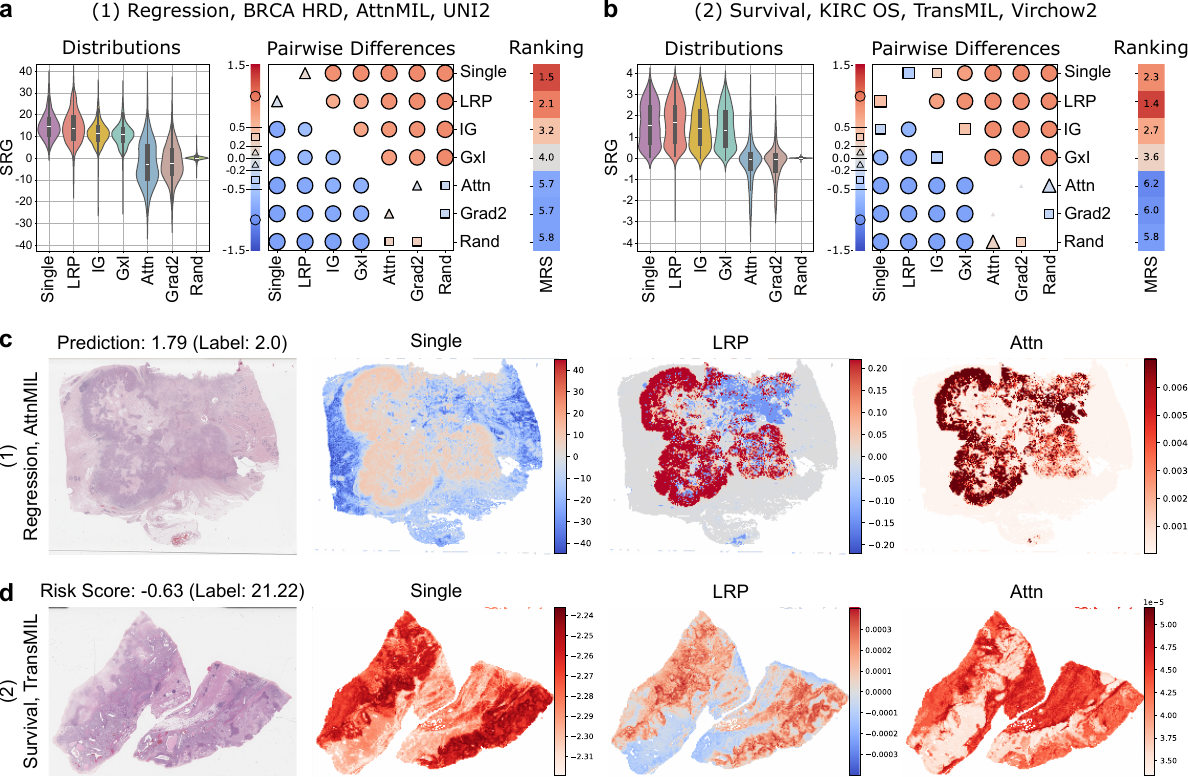}
    \caption{
    \textbf{Comparison of explanation methods for two example settings}.
    \textbf{a}/\textbf{b}~Faithfulness comparison from three complementary perspectives: (\textit{Left}) The distribution of Symmetric Relevance Gain (SRG) faithfulness scores per explanation method computed for each slide in the test set. (\textit{Center}) Wilcoxon's signed-rank test effect sizes comparing SRG scores between explanation methods for each slide in the test set. Colors and marker sizes indicate effect size magnitude as shown on the colorbar. Triangle, square, and circle markers are used to denote the negligible ($< 0.2$), weak to moderate ($0.2-0.5$), and moderate to strong ($\geq 0.5$) effect sizes. These effect size scores assess the magnitude of the faithfulness differences between explanation methods. (\textit{Right}) Mean Rank Score (MRS) per explanation method (vertical axis). \textit{Lower averaged MRS = better position in the faithfulness ranking}.
    \textbf{c}/\textbf{d}~Heatmaps for example slides from the two experimental settings. For LRP, values greater/lower than zero indicate evidence for/against the model prediction (visualized by shades of red/blue). Single and Attention (Attn) have no natural boundary between positive and negative evidence, but only provide an ordering (visualized by shades of red). In the HRD regression task, however, the $0$ value for Single \textit{may} be interpreted as such a boundary; therefore, we visualize this heatmap using a red/blue color scheme as well.
    }
    \label{fig:faithfulness_examples}
\end{figure}

Our heatmap evaluation framework (Section~\ref{sec:methods:evaluating_explanations}) enables a quantitative comparison of explanation methods. Notably, which explanation method performs best depends on many factors, including the dataset and the MIL model. In this section, we demonstrate our methodology for explanation method comparison in two selected applications.
We consider two experimental settings: (1) an AttnMIL regression model with a UNI2 backbone predicting HRD in TCGA BRCA (Spearman correlation: $0.63 \pm 0.03$), and (2) a TransMIL model with Virchow2 backbone predicting overall survival in TCGA KIRC (C-Index: $0.70 \pm 0.03$). First, we visualized the SRG scores for all test set slide heatmaps per explanation method (Fig.~\ref{fig:faithfulness_examples}-a/b,~left). In both settings, the heatmaps of Single, LRP, IG, and G$\times$I achieved SRG scores much higher than random for almost all slides. In contrast, Attn and Grad2 heatmaps were not more faithful than random in many cases. The SRG score distributions of Single, LRP, IG, and G$\times$I had similar appearances, respectively. 

To understand whether the faithfulness differences between the explanation methods were substantial, we analyzed the Wilcoxon's signed rank test's effect sizes between the methods (Figure~\ref{fig:faithfulness_examples}-a/b,~center). In the AttnMIL regression model, Single and LRP performed better than all other methods with a moderate to strong effect size. The differences between those methods, however, were negligible. In the TransMIL survival model, LRP was better than Single (weak to moderate effect size) and IG (moderate to strong effect size). The three methods were better than {G$\times$I, Attn, Grad2, Random} with moderate to strong effect sizes. These results could be reproduced using a robust standardized median based on the median absolute deviation (median/MAD; see Section~\ref{supp:sec:stat_details}).
In addition, to understand the consistency of faithfulness differences across slides, we conducted a ranking-based analysis (Figure~\ref{fig:faithfulness_examples}-a/b,~right). Though SRG score differences were small (see above), Single consistently outperformed LRP in the AttnMIL regression setting (MRS: 1.5 vs.\ 2.1), while LRP consistently outperformed Single and IG in the TransMIL survival setting (MRS: 1.4 vs.\ 2.3).

Consequently, in terms of faithfulness, Single and LRP were the most recommendable methods for the AttnMIL regression model for BRCA HRD with UNI2 backbone, with Single performing slightly better. For the TransMIL overall survival model with Virchow2 backbone for KIRC, LRP performed best with a significant margin over the other explanation methods.

We present example heatmaps for both experimental settings in Figure~\ref{fig:faithfulness_examples}-c/d. For the AttnMIL-based BRCA HRD prediction, both Single and LRP highlighted the same parts of the tumor area to provide evidence in favor of an HRD expression above the clinical reference value of 42 (see Section~\ref{supp:sec:experimental_setup}). LRP additionally separated the remaining slide into a smaller area with contradicting evidence (lower HRD expression) and a larger area of irrelevant features; in the Single heatmap, in contrast, these areas were not distinguishable. We note that, depending on the application, the ability to distinguish irrelevant from contradicting evidence may be an important advantage of LRP over perturbation-based methods. The Attention (Attn) heatmap generally highlighted similar structures as the other heatmaps, but did not make a difference between positive and negative evidence. This exemplifies an important conceptual shortcoming of Attn --- it is impossible to recognize which tissue features are associated with higher and which with lower HRD expression. For the TransMIL-based overall survival prediction in KIRC, Single and LRP also identified similar areas to be evidence of high risk, though with different scaling (color intensities). The Attn heatmap marked considerably different structures, suggesting that it may not be informative in this case.

We note that faithfulness may not be the only selection criterion: LRP is computationally more efficient than IG, while IG is arguably easier to implement and agnostic of the model architecture. LRP additionally divides evidence into positive and negative contributions, which Single cannot do in many cases. Ultimately, our examples show that selecting an explanation method depends on the requirements of the specific application. Yet, faithfulness analysis via patch flipping provides a valuable quantitative evaluation criterion to this end.

\subsection{Comparison of explanation methods across experimental settings}\label{sec:results:faithfulness_stat_comparison}

\begin{figure}[t!]
    \centering
    \includegraphics[width=\linewidth]{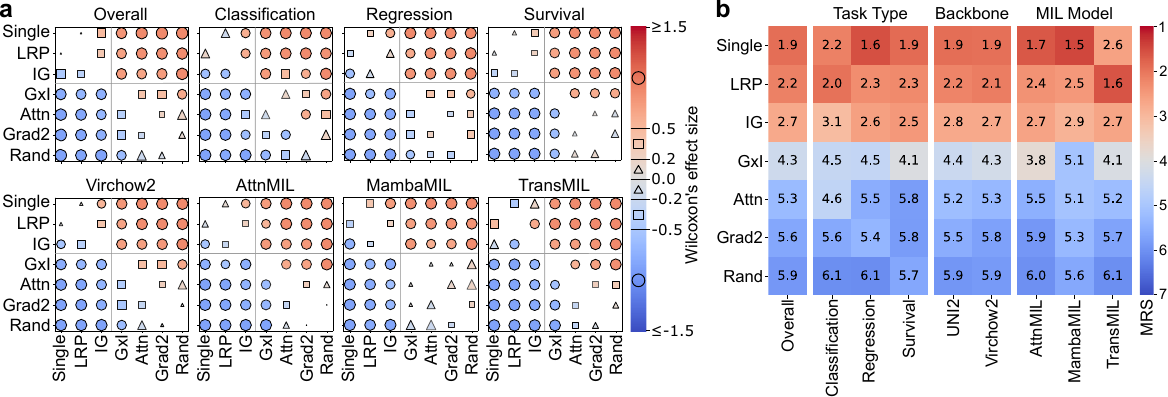}
    \caption{
    \textbf{Comparisons of explanation methods aggregated across experimental settings}.
    \textbf{a}~Pairwise Wilcoxon signed-rank effect sizes comparing SRG scores between explanation methods, averaged over various experimental settings. ``Overall'' considers the mean effect size across all experimental settings, ``Classification'' the mean over all classification experiments, etc. The filling colors and marker sizes indicate effect size magnitude as shown on the colorbar. Triangle, square, and circle markers are used to denote the negligible ($< 0.2$), weak to moderate ($0.2-0.5$), and moderate to strong ($\geq 0.5$) effect sizes. Notably, two consistent groups emerge: {LRP, IG, Single} vs. {Attn, G$\times$I, Grad2, Rand}, with the first group generally outperforming the second. All aggregated comparisons were additionally replicated using a robust median/MAD effect size, and non-significant pairwise comparisons after FDR correction are reported separately in Section~\ref{supp:sec:stat_details}.
    \textbf{b}~Mean Rank Score (MRS) per explanation method (vertical axis), averaged across selected MIL pipeline settings (horizontal axis). \textit{Lower averaged MRS = better position in the faithfulness ranking}.}
    \label{fig:faithfulness_aggregated}
\end{figure}

In addition to the two exemplary experimental settings presented above, we compared the explanation methods across our 60 experimental settings, i.e., ten datasets with three task types (classification, regression, survival), two patch encoder backbones (UNI2, Virchow2), and three aggregation models (AttnMIL, MambaMIL, TransMIL). Specifically, we aggregated the faithfulness results from different perspectives, enabling us to identify overall trends and recommendations.

First, we analyzed the aggregated pairwise Wilcoxon's signed rank test's effect sizes between the explanation methods (Figure~\ref{fig:faithfulness_aggregated}-a).
Figure~\ref{fig:faithfulness_aggregated}-a demonstrates a consistent grouping of methods: {LRP, IG, Single} versus {Attn, G$\times$I, Grad2, Rand}. The first group outperforms the second with strong effect sizes, while within-group differences are only moderate to weak, indicating broadly similar levels of faithfulness. This observation is replicable using a standardized median as a statistical measure of paired differences (see Section~\ref{supp:sec:stat_details}). 
Importantly, Attention heatmaps are found almost always substantially less faithful than LRP, IG, and Single, with strong negative effect sizes. This pattern is consistent across almost all MIL models, prediction tasks, and backbones. 
Notably, the Random (Rand) baseline clusters with Grad2, G$\times$I, and Attn in all cases, raising questions about their added value in those contexts.

In classification tasks, IG performs worse than LRP and Single, whereas Attention appears more competitive than in other settings. 
Similarly, in AttnMIL, G$\times$I gets closer to LRP (without a statistical significance). This suggests that for simpler tasks and architectures, simpler explanation methods may approach the performance of more sophisticated ones. 
In regression tasks and MambaMIL, Single is clearly the best-performing method, differing with moderate to strong effect sizes from LRP and IG. In contrast, in survival prediction, Single shows weak and moderate differences to LRP and IG, respectively. For the Transformer-based architecture, LRP is the best-performing method, with moderate to strong differences to Single and IG.

Second, we analyzed the aggregated rankings of the explanation methods (Figure~\ref{fig:faithfulness_aggregated}-b), which identify consistent faithfulness differences between methods, even if they are small in effect size.
Across all experimental settings (``Overall''), Single performed best on average (1.8 MRS), closely followed by LRP (2.1 MRS). IG still performed competitively (2.8 MRS). G$\times$I, Attention, and Grad2 were clearly worse (4.4-5.9 MRS) than the other methods, which is in accordance with the effect size analysis presented previously.
Similar rankings were observed for the foundation model backbones UNI2 and Virchow2.
In classification tasks, Single and LRP achieved comparable faithfulness ranks, while IG was considerably worse. In regression and survival tasks, however, the Single method outperformed LRP and IG.
The choice of the MIL model had the most notable influence on the ranking of the explanation methods. While Single excelled for AttnMIL and MambaMIL, LRP was consistently the best method for TransMIL.

Overall, our large-scale faithfulness analysis reinforced the notion that the best explanation method depends on the experimental setting. At the same time, we observed that Single, LRP, and IG were considerably and consistently more faithful than Attention and other methods. Between Single, LRP, and IG, faithfulness differences were often (but not always) negligible.

\section{Use case 1: Validating heatmaps with biological ground-truth data} \label{sec:hest_biological_val}

In the previous section, we showed that explanation methods such as Single, LRP, and IG produce heatmaps that are consistently and substantially more faithful to MIL model predictions than other methods, particularly attention heatmaps. In this section, we investigate whether these heatmaps also convey biologically meaningful information. This question is also relevant from a model validation perspective: it remains unclear how much signal can be predicted from histology alone, making it essential to exclude shortcut learning. To this end, we assess whether attribution heatmaps exhibit spatial concordance with independently measured biological ground truth rather than reflecting spurious associations.
As a \textit{proof-of-concept}, we designed a novel validation experiment to assess whether explanation methods can map (bulk) molecular biomarker predictions to spatial signals. We used a dataset of H\&E-stained whole slide images paired with spatial transcriptomics (ST) data, and trained a MIL regression model to predict pseudo-bulk gene expression. We hypothesized that the attribution heatmaps of the MIL model would reflect the spatial distribution of gene expression, as the MIL model aggregates spatial evidence contributing to the slide-level gene expression prediction. We evaluated this hypothesis by measuring correlations between heatmap scores and ST ground truth expression.

\subsection{Experimental setup}

\begin{figure}[!t]
    \centering
    \includegraphics[width=\linewidth]{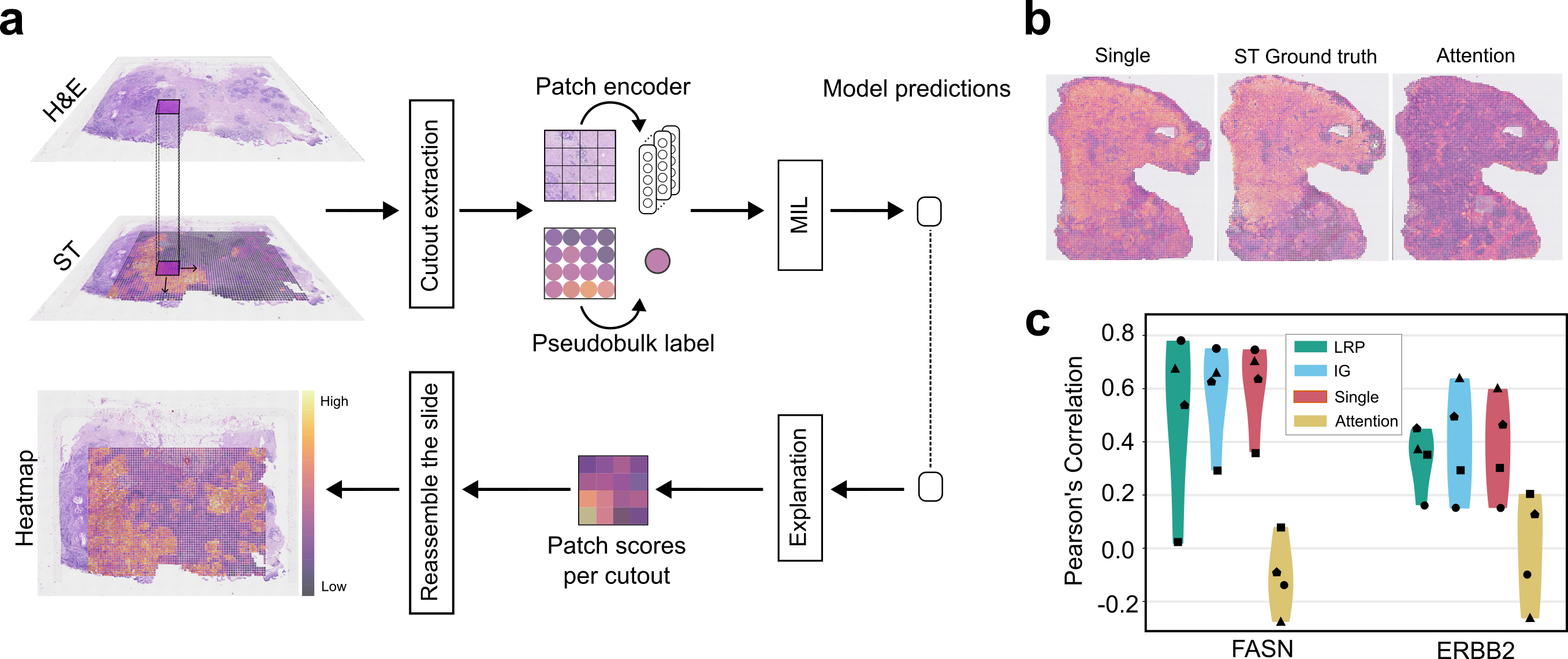}
    \caption{
    \textbf{Proof-of-concept for spatial mapping of bulk gene expression using explanation heatmaps of a regression MIL model}.
    \textbf{a} Experimental pipeline. Whole-slide H\&E images are subdivided into cutouts, from which patch-level feature vectors are extracted. For each cutout, pseudo-bulk gene expression targets are derived by aggregating spatial transcriptomics (ST) measurements spatially aligned to the same region. Model explanation methods are then applied to obtain patch-level relevance scores, which are then realigned to build a WSI-level heatmap and compared against the original spatial gene expression.
    \textbf{b}~Example heatmaps for gene FASN on a slide, shown alongside the corresponding ST ground truth. Heatmaps indicate the patch-level model-attributed relevance for predicting bulk expression.
    \textbf{c}~Pearson correlation coefficients between patch scores derived from explanation heatmaps and the corresponding ST ground truth. Results are shown for four explanation methods: Layer-wise Relevance Propagation (LRP), Integrated Gradients (IG), Single, and Attention. Individual slides are indicated by distinct symbols on the violin plots.
    }
    \label{fig:HEST_overview}
\end{figure}

Figure~\ref{fig:HEST_overview}-a shows the experimental pipeline. We used a subset of four WSIs from four human breast cancer cases in the HEST-1k dataset \citep{jaume2024hest}. In HEST-1k, ST measurements are spatially registered to pre-extracted H\&E patches of the same tissue, yielding a gene expression vector per patch. Due to the limited number of slides, we augmented the data by subdividing each slide into multiple non-overlapping regions, referred to as \textit{cutouts}, each covering up to $10 \times 10$ patch area on the tesselated slide. As prediction targets, we selected two exemplary genes (FASN, ERBB2) from the top 50 highly variable genes. Selection criteria were: (1) non-zero expression in at least 10\% of ST spots, and (2) feasibility of training a well-performing pseudo-bulk prediction model. Pseudo-bulk expression for each cutout was obtained by normalizing spot-level expression values and aggregating them via sum pooling across patches within the cutout. Each cutout was treated as a bag of patches, with the pseudo-bulk expression serving as the bag-level label. For evaluation, we trained models on cutouts of three slides and tested on the cutouts of the remaining slide.

We trained TransMIL regression models with a Virchow2 backbone to predict the pseudo-bulk gene expression from the H\&E patches inside the cutout (details on training and model selection in Section~\ref{supp:sec:HEST_training_details}). For the cutouts of the held-out slide, we generated heatmaps using LRP, IG, Single, and Attention. Afterwards, cutout-level heatmaps were reassembled into full-slide heatmaps. We quantified the biological relevance of the explanation heatmaps by computing the Pearson correlations between the patch-level heatmap scores and their corresponding ground-truth gene expression values from the ST data. Note that the original spot-level ST expressions were never seen by the MIL model: the model was trained to predict the pseudo-bulk expressions.

\subsection{Results}

We evaluated regression performance by correlating cutout-level predictions with pseudo-bulk labels on the validation set. The average model performance across the cross-validation folds were Pearson correlations of 0.80 for FASN and 0.75 for ERBB2. These results confirm that (pseudo-)bulk gene expressions can be predicted from H\&E regions with reasonable performance.

Figure~\ref{fig:HEST_overview}-b shows example heatmaps for FASN alongside its corresponding ST ground truth. The Attention heatmap is noisy and fails to highlight regions of high FASN expression. This observation does not contradict the overall model performance, as attention values are known to emphasize the features that are necessary but not sufficient for the final model prediction \citep{javed2022additivemil}. In the same panel, the heatmap depicting the Single attributions closely matches the spatial distribution of ST measurements, capturing fine-grained regional differences across the slide. This suggests that the MIL model learned biologically relevant features corresponding to FASN expression and that Single effectively exposes them. Similar patterns are observed for most other (but not all) slides and explanation methods (see Figures~\ref{supp:fig:HEST_all_heatmaps_ERBB2}, \ref{supp:fig:HEST_all_heatmaps_FASN}).

Figure \ref{fig:HEST_overview}-c shows the Pearson correlations for all slides, explanation methods, and genes. Across both genes, Attention heatmaps consistently exhibit lower Pearson correlations with ST ground truth than LRP, IG, and Single. For FASN, LRP, IG, and Single achieve strong Pearson correlations (0.54-0.78) on three of four slides; on the remaining slide, IG and Single achieve moderate correlations, while LRP does not. For ERBB2, two slides show medium to high correlations (0.37-0.64), one slide shows moderate alignment (0.29-0.35), and one exhibits weak correlations (0.15-0.16).

Overall, these experiments showcase that more faithful heatmaps can provide biologically meaningful spatial signals in complex molecular biomarker prediction tasks.
We anticipate useful potential implications: First, assessing MIL heatmaps through matched spatial transcriptomics could enable a novel approach to validating molecular biomarker prediction MIL models. So far, such MIL models could only be evaluated based on their prediction performance; adding a spatial transcriptomics assessment before clinical deployment could increase trust. Second, after validation, heatmaps could be created for large data cohorts (for which ST is too expensive) to guide clinicians or researchers in systematically identifying tissue regions of high or low gene expression.


\section{Use case 2: Systematic discovery of distinct model strategies} \label{sec:cell_vit_exploration}

\begin{figure}[!t]
    \centering
    \includegraphics[width=\linewidth]{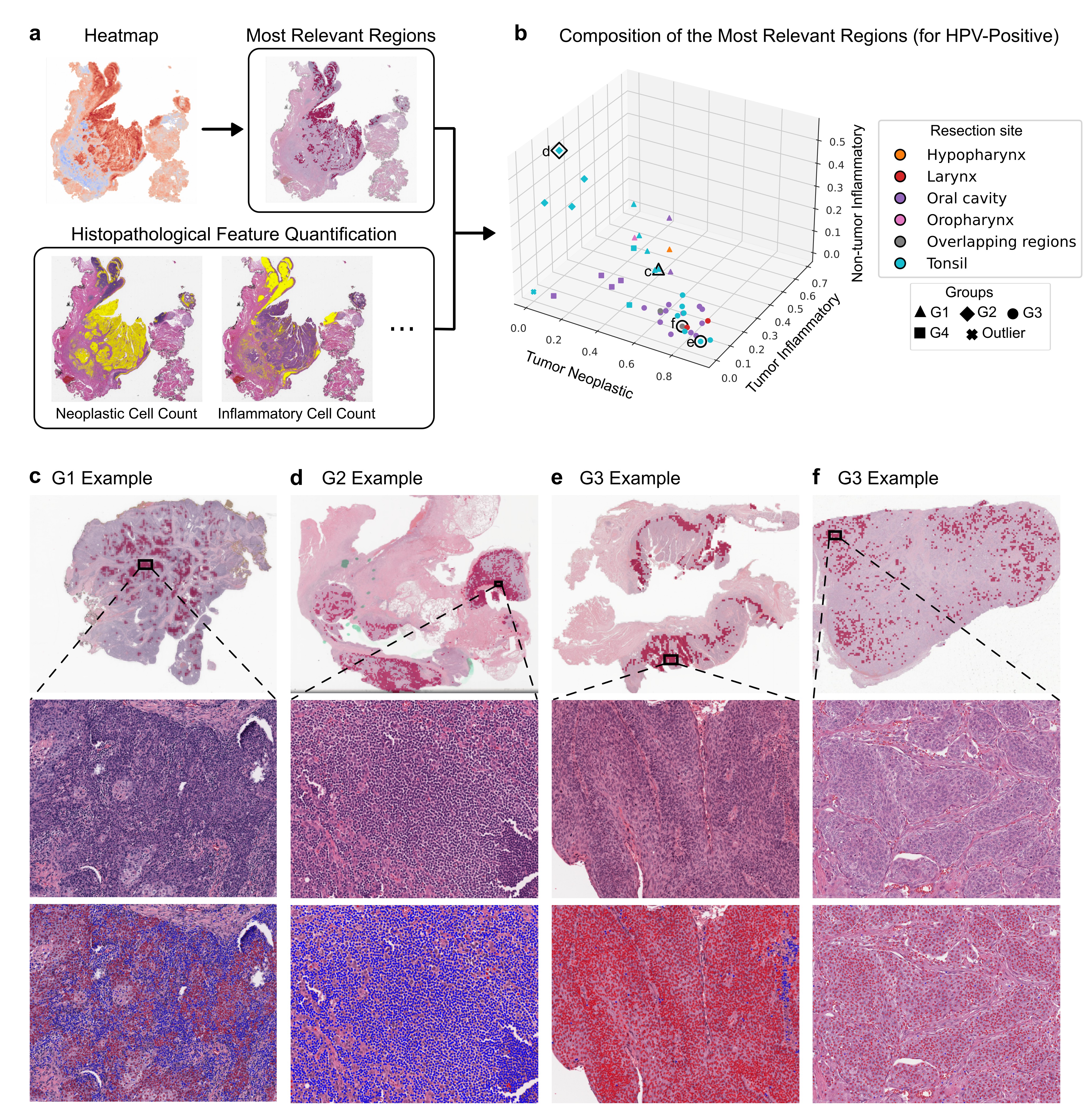}
    \caption{
    \textbf{Histological analysis of HPV heatmaps for discovery}.
    \textbf{a}~We extract the most relevant heatmap regions (=top 10\% LRP patch scores) and characterize them by computing interpretable and quantifiable histological features.
    \textbf{b}~The cell composition of the most relevant regions determined via CellViT++ \citep{horst2025cellvitpp} and tissue compartment segmentation exposes clusters among HPV-positive slides. See also Figure~\ref{supp:fig:scatter_matrix} for a 2D scatter matrix of the five most relevant features. The samples in panels c-f are marked. 
    \textbf{c-f}~Example heatmaps and relevant histological features for HPV-positive prediction, consisting of a thumbnail of the most relevant regions (top row) and a selected zoomed-in slide region (middle row) with CellViT-annotated neoplastic (red) and inflammatory cells (blue) (bottom row).
    }
    \label{fig:hpv_analysis}
\end{figure}

After validating the potential of Single, LRP, and IG to produce faithful and biologically meaningful heatmaps, we now showcase how such heatmaps can be combined with human-interpretable features to identify and group connections between input patterns and prediction targets. Traditionally, heatmap analysis relies exclusively on qualitative visual inspection by a pathologist, which is laborious and prone to confirmation bias. Instead, we present a quantitative approach to complement the visual inspection (similar to \cite{markey2025spatial, mokhtari24interpretable, rodrigues2025imilia}). Specifically, we extract the most relevant heatmap regions and characterize them by computing histological features such as cell counts, nuclei sizes, or tissue organization (Figure~\ref{fig:hpv_analysis}-a). This translates heatmaps into quantitative, human-interpretable histological profiles associated with the predicted target, which may point to biological mechanisms or confounding correlations.

\subsection{Experimental setup}

We revisit HPV prediction in head and neck squamous cell carcinoma (HNSCC), using a TransMIL model with Virchow2 backbone (AUROC $0.90 \pm 0.04$, see Section~\ref{supp:sec:experimental_setup:datasets} and Figure~\ref{fig:model_dataset_figure}-a).
This task is particularly well-suited for discovery: although certain histological features of HPV-positivity are known (i.e., smaller tumor cells with hyperchromatic nuclei and dense lymphocyte infiltration), the diagnostic gold standard remains immunohistochemistry (p16) and/or molecular testing \citep{bilal2023aggregation, campbell2018tcgascc}. This suggests that additional, yet unknown histological indicators may exist. 

For a basic tissue characterization, we first segmented each slide into tumor, border, and non-tumor compartments using a patch classification model from \cite{hense2025pathways} (see Section~\ref{sec:supp:use_case_2_details} for details). We then applied CellViT++ \citep{horst2024cellvit, horst2025cellvitpp} to segment all cells of the slide and classify them as Neoplastic, Inflammatory, Connective, Epithelial, or Dead. 
We then defined the region of interest (ROI) as the patches with the highest explanation scores (top 10\% LRP scores); i.e., regions most relevant to HPV-positivity. We refer to this ROI as the \textbf{High-LRP 10\% (HL10)} region.
Consequently, for each cell type-compartment combination, we computed the proportion of cells within the ROI. This approach yields a simple morphological profile per slide that quantifies the visual features the model uses as evidence for HPV-positivity.

\subsection{Results}

We first examined how much each histological feature was relevant to HPV-positivity according to the model's prediction strategy (i.e., what is the share of each cell type-compartment combination in HL10, see Figure~\ref{fig:feature_relevance}). Across all HPV-positive samples in the HNSCC dataset, the HL10 region was predominantly composed of neoplastic cells in the tumor compartment (mean $\pm$ std.: $51.42\% \pm 27.69\%$), inflammatory cells in the tumor compartment ($17.78\% \pm 16.18 \%$), inflammatory cells in the nontumor compartment ($4.94 \% \pm 12.42 \%$) and connective cells in the tumor ($9.46\% \pm 6.5\%$) and nontumor compartment ($3.77\% \pm 7.9\%$). The other cell types and the border region contributed substantially smaller shares. In order to understand different model strategies, we clustered the HPV-positive slides using the extracted morphological profile into five clusters (Agglomerative Clustering, number of clusters selected based on the maximum average Silhouette Coefficient). We denote the first four clusters as groups \textbf{G1}, \textbf{G2}, \textbf{G3}, \textbf{G4}), suggesting that the model relied on different histological markers that it associates with HPV-positivity. The fifth cluster was a single-member cluster, which we marked as an outlier.
We then compared the HL10-based histological profiles across HPV-positive slides to identify potentially different model prediction strategies. The share of connective cells within the HL10 region was consistently low and mostly homogeneous across HPV positive samples (see Figure~\ref{supp:fig:scatter_matrix}). In contrast, the relative amounts of neoplastic and inflammatory cells revealed distinct patterns of cell profiles (Figures~\ref{fig:hpv_analysis}-b and \ref{supp:fig:scatter_matrix}). 

\paragraph{\textbf{G1}} 
In group G1, the HL10 region was characterized by an exceptionally high proportion of intratumoral inflammatory cells (HL10 share: $43.83 \pm 14.06\%$), resulting in a histological profile distinct from the other samples. A pathological inspection of the heatmaps by AM confirmed pronounced lymphocyte infiltration in tumor areas for samples in this group, a known marker of HPV-positivity (example in Figure~\ref{fig:hpv_analysis}-c).
\paragraph{\textbf{G2}} 
In group G2, the HL10 regions showed a high share of inflammatory cells in the nontumor compartment. Notably, all slides in this group originated from the tonsil (Figure~\ref{fig:hpv_analysis}-b). This may indicate that the model recognized an HPV infection in this group by learning the primary anatomical side of tumors harboring an HPV infection (i.e., tonsil and oropharynx) (example in Figure~\ref{fig:hpv_analysis}-d).
\paragraph{\textbf{G3}} 
In contrast, group G3 was dominated by neoplastic cells ($73.60\% \pm 10.24 \%$) combined with a low inflammatory cell content ($11.01\% \pm 6.26\%$) in the tumor compartment. Many of the slides in this group were correctly classified as HPV-positive (Figure~\ref{fig:hpv_analysis}-b). Interestingly, for five HPV-positive slides in this group, pathological inspection of the whole slide did not identify any typical signs of HPV-positivity. Instead, we observed typical characteristics of HPV-negative tumors, such as lower to no lymphocyte infiltration in the tumor areas as well as keratinization (examples in Figure~\ref{fig:hpv_analysis}-e,f). \textit{This could suggest that the model identified histological patterns predictive of HPV-positivity that were not recognized by the pathologist.}
\paragraph{\textbf{G4}} 
In group G4, none of the three investigated features were strongly pronounced within HL10, suggesting that additional analyses or further features would be needed to develop a better understanding of the model strategy for these samples.

Overall, we demonstrated a proof-of-concept that overlaying well-performing heat-\linebreak[1]maps with quantifiable morphological features enables the identification of different model strategies and supports hypothesis generation about biological associations to the prediction target. Furthermore, this XAI-driven approach can help identify model failure modes. We anticipate that such an analysis can yield significantly more nuanced and robust results when using more fine-grained morphological features (such as \cite{markey2025spatial}) or larger data cohorts with a specific discovery target.


\section{Discussion} \label{sec:conclusion}

MIL heatmaps are widely used in computational pathology for model validation, hypothesis generation, and biological discovery. They are often presented as evidence that a model focuses on morphologically meaningful regions, and are considered a standard component of end-to-end MIL pipelines for whole-slide image prediction \citep{elnahhas2025weakly}. Despite their central role, heatmap evaluation is commonly based on qualitative expert inspection, and different studies adopt heterogeneous explanation methods without a principled rationale for method selection. As a result, there is a lack of established guidelines for choosing or validating explanation techniques in MIL settings. Prior work has argued that clinically meaningful explanations must satisfy quantitative criteria of truthfulness and informative plausibility when interpreting AI models on medical images \citep{jin23guidelinesxai}. However, systematic methods for assessing heatmap quality or selecting appropriate explanation techniques in histopathology MIL remain scarce.

\textit{We posit that the first step in evaluating heatmaps should move beyond expert visual assessment and focus on whether a heatmap technically reflects the model’s output.} To this end, we propose a method-agnostic, data-driven heatmap evaluation framework that operates without additional labels. 
Through extensive experiments, we apply this framework to benchmark six explanation methods across a first-of-its-kind, large-scale experimental design, including major prediction task types (classification, regression, survival), backbones (Virchow2, UNI2), and MIL model architectures (Attention, Transformer, Mamba). The results show that task type and model architecture are the primary factors in selecting the best explanation method, and that attention heatmaps are rarely competitive with input perturbation (``Single''), LRP, and IG.
Finally, we demonstrate that well-performing MIL explanation methods enable new applications and analyses. In two novel proof-of-concept use cases, we employed heatmaps for validating bulk biomarker prediction models through spatial transcriptomics and to identify sample groupings with distinct prediction strategies for generating biological hypotheses.

We highlight that our analysis of pairwise effect sizes and faithfulness rankings reveals a robust grouping of explanation methods: LRP, IG, and Single consistently outperform Attention, G$\times$I, Grad2 across MIL models, task types, and backbones. Attention heatmaps, in particular, do not differ much from the random baseline in multiple cases. This is a notable observation given that the Attention heatmap is used very frequently as an ``explanation'' for black-box MIL models. Importantly, these results remained stable under two different effect size definitions and when examining the statistical significance of the pairwise statistical tests.
From our observations, we derive the following practical recommendations:
\begin{enumerate}
    \item For Transformer-based MIL models, LRP is generally the most appropriate explanation method.
    \item Single is likely to be the most faithful method for Attention- and Mamba-based MIL models, particularly in regression and survival tasks.
    \item IG may serve as an attractive alternative in cases where the two better methods, Single and LRP, come with high implementation effort. IG is available from toolboxes like Captum \citep{kokhlikyan2020captum} and can be easily adapted to MIL models.
\end{enumerate}
These recommendations provide a useful starting point for narrowing the search space of explanation methods. However, the optimal choice may still depend on the specific application. Therefore, we recommend systematically evaluating multiple explanation methods for each use case.

We anticipate that our work will support the community in generating more reliable and interpretable heatmaps and structuring heatmap selection decision-making, thereby expanding the possibilities for model validation, contributing to overcoming the currently limited clinical adoption of machine learning in digital pathology. 
This is particularly relevant for validating promising biomarker prediction models (e.g., \cite{campanella2025real, wagner2023transformer}), a task that has traditionally required assembling large validation cohorts.
Furthermore, heatmaps of bulk omics prediction models may enable cheap spatial analyses for omics data at scale. While recent work has shown the great potential of ML-generated virtual spatial omics for discovery \citep{klockner2025virtual_staining, li2026protein, schmauch2025omics, valanarasu2025gigatime}, these approaches require training data of ground-truth spatial omics data. In contrast, our heatmap-based approach only requires training a MIL model to predict bulk omics targets, allowing for spatial analyses where no spatial omics data are available. We note that in concurrent work, a heatmap-based spatial mapping of pathway activities in head and neck tumor slides enabled the identification of a novel phenotype associated with disease recurrence \citep{hense2025pathways}.
Finally, improved heatmaps may enable a novel way to discover targetable morphological patterns in patient subgroups associated with clinical endpoints, potentially informing drug discovery research and improving patient outcomes.


Several aspects of this work offer opportunities for further development.
First, the explanation and evaluation methods considered here capture only first-order contributions, and therefore cannot explicitly model feature interactions that the MIL model may exploit for prediction. Extending higher-order explanation techniques \citep{eberle2020building, schnake2021higher, schnake2025symbolic} to the MIL setting could reveal previously unknown biological interactions among morphological features.
Second, the XAI community has proposed further desirable properties for explanations beyond faithfulness, such as robustness to perturbations or contrastivity relative to targets that did not occur \citep{hedstrom2023quantus, jin23guidelinesxai, klein2024navigating, nauta2023xai_eval}. In the context of MIL heatmaps, it would be particularly valuable to additionally assess whether the magnitude of explanation scores reflects the actual influence of each patch on the prediction in terms of the output unit, such as HRD score or survival risk \citep{ancona2018towards}.
Third, we did not assess inherently interpretable MIL models such as additive MIL \citep{javed2022additivemil}. While these methods can provide insightful heatmaps \citep{markey2025spatial}, the post-hoc explanation methods examined in this work are more versatile as they do not constrain the underlying model architecture, which may be crucial for achieving competitive performance in practical applications.
Finally, although this study focuses on histopathology, the explanation and evaluation approaches directly transfer to other domains and use cases, particularly in radiology, single-cell and spatial molecular data, and multimodal fusion models, where explanations could uncover modality-specific and shared patterns associated with patient outcome.

\section*{Code and data availability}
Our code for reproducing our results and implementation of the used methods is made publicly accessible at \url{https://github.com/bifold-pathomics/xMIL/tree/xmil-journal}. Datasets used in this work are public datasets described and cited in Section~\ref{supp:sec:experimental_setup:datasets}.

\section*{Declaration of competing interest}
F.K. and K.R.M. are co-founders and J.D. is an employee of Aignostics GmbH, which develops AI algorithms for pathology. The other authors have no competing interests to declare that are relevant to the content of this article.

\section*{CRediT authorship contribution statement}
\textbf{Mina Jamshidi Idaji \& Julius Hense:} Conceptualization, Methodology, Software, Validation, Data Curation, Investigation, Formal analysis, Visualization, Project administration, Writing – Original Draft, Supervision.
\textbf{Tom Neuh{\"a}user:} Methodology, Software, Validation, Investigation, Writing – Original Draft.
\textbf{Augustin Krause:} Methodology, Software, Validation, Investigation, Data Curation, Visualization, Writing - Original Draft.
\textbf{Oliver Eberle \& Thomas Schnake:} Conceptualization, Methodology.
\textbf{Yanqing Luo:} Software, Investigation, Data Curation.
\textbf{Farnoush Rezaei Jafari \& Reza Vahidimajd \& Laure Ciernik \& Jonas Dippel:} Software.
\textbf{Christoph Walz:} Investigation.
\textbf{Andreas Mock:} Data Curation, Investigation, Writing - Original Draft.
\textbf{Klaus-Robert M{\"u}ller:} Supervision, Funding acquisition.
\textbf{All authors:} Writing – Review \& Editing.

\section*{Acknowledgements}
We thank Simon Leszek for the fruitful discussions on explaining regression models. We thank Magdalena Gippert for the discussions on statistical aspects of the study. We also thank David Drexlin for his feedback on the manuscript.
The results shown here are in part based upon data generated by the TCGA Research Network: https://www.cancer.gov/tcga. This work was in part supported by the German Ministry for Education and Research (BMBF) under Grants 01IS14013A-E, 01GQ1115, 01GQ0850, 01IS18025A, 031L0207D, 01IS18037A, and BIFOLD24B and by the Institute of Information \& Communications Technology Planning \& Evaluation (IITP) grants funded by the Korea government (MSIT) (No. 2019-0-00079, Artificial Intelligence Graduate School Program, Korea University and No. 2022-0-00984, Development of Artificial Intelligence Technology for Personalized Plug-and-Play Explanation and Verification of Explanation). Thomas Schnake is a postdoctoral fellow at the University of Toronto in the Eric and Wendy Schmidt AI in Science Postdoctoral Fellowship Program, a program of Schmidt Sciences. 


\FloatBarrier
\newpage

\appendix

\section*{\LARGE Supplementary}

This supplementary material provides methodological details, experimental specifications, and additional analyses supporting the main manuscript. It includes: 
(\ref{supp:sec:related_work}) An extended review of multiple instance learning (MIL) architectures and interpretability methods; 
(\ref{supp:sec:survival_preliminaries}) Detailed formulations related to the survival models implemented in the manuscript; 
(\ref{supp:sec:exp_methods}) Comprehensive descriptions and mathematical derivations of all explanation methods, including LRP rules; 
(\ref{supp:sec:experimental_setup}) Full experimental settings covering datasets, preprocessing, foundation models, training protocols, hyperparameter grids, and performance results; 
(\ref{supp:sec:stat_details}) Statistical analyses details and results related to Section~\ref{sec:results:faithfulness_stat_comparison} of the main paper; 
(\ref{supp:sec:HEST_training_details}) Additional experiments and figures for Section~\ref{sec:hest_biological_val} of the main manuscript; and 
(\ref{sec:supp:use_case_2_details}) Details of the analyses related to Section~\ref{sec:cell_vit_exploration} of the main paper. 
Some parts of this supplementary are shared with the appendix of \cite{hense2024xmil}. 

\vspace{5mm}

\section{Related work} \label{supp:sec:related_work}

\subsection{MIL model architectures}

In this section, we review popular multiple instance learning model architectures. We do not provide an exhaustive review due to the myriad of available literature, but instead focus on representative examples.

Traditional multiple instance learning models in histopathology relied on computing pooling patch-level prediction scores. \cite{ilse2018attentionmil} introduced attention-based pooling, allowing a model to learn slide representations by dynamically weighting instances. This architecture has been extended into various directions, e.g., by adding multi-class attention \citep{lu2021clam}, multi-scale mechanisms \citep{li2021dualstreammil}, low-rank attention \citep{xiang2023ilra}, kernel-based self-attention \citep{kernelmil2021}, patch representation clustering \citep{sharma2021clustermil}, and changes to the training algorithm \citep{schirris22deepsmile, yang2022remixmil, zhang2022dtfdmil}.

Moving towards Transformer-based MIL models, \cite{shao2021transmil} proposed the TransMIL architecture, which outperformed previous methods by a large margin. Again, several other Transformer-based models have been proposed, e.g., with restrictions in the self-attention mechanism for efficiency and smoothness \citep{castro2024sm, fourkioti2024camil, li2024longmil}, a regional Transformer for feature re-embedding \citep{tang2024rrt}, or removal of positional encodings \citep{wagner2023transformer, shao2025do}.
Recent works propose to adopt state-space models to multiple instance learning \citep{fillioux2023s4mil}. A popular representative is MambaMIL \citep{yang2024mambamil}, which models patch representations as an input sequence for a Mamba layer \citep{gu2024mamba}.

Alternative MIL architectures are built on graphs to model spatial relationships between patches \citep{bazargani24graph, lee2022derivation, li2024wikg}. Other works investigate how to conduct zero- or few-shot learning using language-alignment in MIL \citep{lu23mizero, meseguer26mil_adapter}.
Various studies have benchmarked MIL model architectures in different contexts \citep{bilal2023aggregation, ghaffari22benchmarking, shao2025do, xu25milfm}. Yet, which architecture performs best arguably depends on various factors, including dataset, prediction task, and feature extractor.

\subsection{MIL model interpretability}

Interpretability or explainability is an omnipresent topic in multiple instance learning and histopathology. It is often approached through patch or instance attributions. The most common way to obtain such attributions is to extract model-internal attention scores to create attention heatmaps. This was initially introduced by \cite{ilse2018attentionmil} for the Attention MIL architecture and later extended to other MIL approaches such as multi-class attention \citep{lu2021clam} and Transformers \citep{abnar-zuidema-2020-quantifying, shao2021transmil}.

Various methods have been proposed to improve the original attention heatmaps. Some suggest modifications in the model architecture or training process to enhance the quality of the attention heatmaps \citep{castro2024sm, DENG2024103124, fourkioti2024camil, hu2025self, SHI2024103294} or combine attention with additional model outputs such as patch prediction scores \citep{bonnaffe2023beyond, cai2025attrimil, sharma2025method, wagner2023transformer}. Other authors highlight the shortcomings of attention heatmaps and instead resort to other XAI-based attribution methods. A popular choice is the gradient-based Integrated Gradients (IG) approach \citep{lee2022derivation, li2023vision, song2024analysis, vu2025contrastiveig}. Recent works adopt propagation-based methods such as Layer-wise Relevance Propagation (LRP) \citep{hense2024xmil, hense2025pathways, sadafi2023pixelmil}. Furthermore, \cite{early2022shapmil} proposed bag perturbation mechanisms, i.e., deriving patch attribution scores from observing changes in the model output when patches are added or dropped, similar to SHAP \citep{lundberg2017shap}. \cite{javed2022additivemil} introduced Additive MIL, an inherently interpretable additive reformulation of the original Attention MIL architecture. Similarly, some authors developed model architectures that simultaneously produce patch and slide predictions, e.g., to obtain a disease area segmentation in addition to the disease classification \citep{gao2025smmile, huang2026geometric}.

Heatmaps are frequently used for model validation, i.e., to confirm that focus areas of the MIL model concur with the expectations of pathologists. Examples include applications in disease detection and classification \citep{lipkova2022deep, lu2021ai, calderaro2023deep, jiang2024transformer, kurata2025multiple, tauqeer2025detection}, biomarker prediction \citep{Graziani2022reg, wagner2023transformer}, and survival prediction \citep{chen2022pan, song2024analysis}. \cite{bouzid2024barretts} overlaid attention heatmaps with spatially registered IHC stains to quantitatively confirm that their MIL model for screening Barrett's esophagus focuses on the presence of goblet cells, a known indication of the disease.

Beyond model validation, MIL heatmaps have been used for the discovery of novel histological markers associated with relevant clinical endpoints. \cite{mokhtari24interpretable} and \cite{rodrigues2025imilia} connected attention heatmaps and patch prediction scores with interpretable cell classification features to identify histological markers of inflammatory bowel diseases (IBDs). \cite{lee2022derivation} employed IG heatmaps to find histopathological prognostic markers in clear cell renal cell carcinoma, and \cite{li2023vision} to detect diagnostic histopathological features of primary brain tumor subtypes. \cite{markey2025spatial} trained an Additive MIL model to predict TGF$\beta$-CAF, a gene expression signature associated with immunotherapy treatment response, and overlaid the heatmaps with 297 human-interpretable tissue features to discover morphological markers of this transcription mechanism.

Few studies focus on evaluating and comparing heatmaps. \cite{jang2024multiple} and \cite{gao2025smmile} provided theoretical analyses of the potential of heatmaps to classify patches under traditional MIL assumptions \citep{DIETTERICH199731, ilse2018attentionmil, NIPS1997_82965d4e}, and \cite{gao2025smmile} and \citep{huang2026geometric} conducted experimental assessments in this regard. \cite{javed2022additivemil} and \cite{AFONSO2024100403} further compared Attention MIL and Additive MIL on tumor detection and TP53 biomarker prediction tasks. \cite{sadafi2023pixelmil} assessed gradient- and propagation-based explanation methods in the context of a pixel-level attribution task.

Our work substantially differs from the existing literature:
\begin{enumerate}
    \item 
    First, most existing application works use a chosen explanation method without validating the quality of the heatmaps. In contrast, we emphasize the evaluation step and propose a framework to assess and compare heatmap quality to select the explanation method best-suited for the specific application at hand. 
    \item 
    Second, while most papers focus on a single selected explanation method in the context of a specific experimental configuration with a fixed MIL architecture and a chosen dataset, we consider six MIL explanation methods across a wide range of experimental settings, e.g., backbones, MIL model architectures, and prediction tasks. This allows us to derive general insights beyond the limits of a specific application.
\end{enumerate}

\section{Preliminaries on survival modeling} \label{supp:sec:survival_preliminaries}

\subsection{Details on survival modeling} \label{supp:sec:survival_modeling_details}  

One of the most popular modeling approaches is the Cox proportional hazards (CPH) model \citep{katzman2018deepsurv, Prentice1992}, where the patient hazard is modeled as an exponential linear function of the outcome of a neural network as $h(t|\mathbf{x}) = h_0(t)\exp(f(\mathbf{x}))$ with $f(\mathbf{x}) \in \mathbb{R}$ being the regression-like output of a neural network representing an overall patient log-risk score, and $h_0(t)$ a baseline hazard. The neural network is commonly trained with the Cox partial log-likelihood, which pushes the model to assign hazard scores that correctly rank the patients according to their survival time. To achieve this, it requires comparing multiple patient samples in a mini-batch \citep{chen21mcat, katzman2018deepsurv}.
When training MIL models on patient data with a typically large number of patches per slide, this approach is problematic, as the mini-batch size is substantially limited due to memory and computational restrictions. In practice, Transformer-based MIL models are frequently trained with batch size 1 \citep{chen21mcat, wagner2023transformer, shao2025do}.
Therefore, we consider a popular alternative approach of discrete-time survival models \citep{chen21mcat, zadeh2020bias}. In this setting, the survival period is divided into a fixed number of non-overlapping intervals $T_{1}=[t_{0}, t_{1}), \ldots, T_{K}=[t_{K-1}, t_{K})$ (we use $K=4$ intervals and hereafter refer to this number). Each patient’s event time is assigned to a discrete interval. The aggregated bag embedding is first mapped by a linear layer with sigmoid activation to the hazard probabilities $\{h_{k}=\sigma(f(X)_{k})\}_k \subset (0,1)$ representing the discrete hazard probabilities for the pre-defined time intervals indexed with $k$. These logits are then passed through a survival head, which computes the final risk score $r$ by combining the hazard probabilities and survival function as
\begin{align}
    S_k = \prod_{j=1}^{k} (1 - h_{j}) \\
    r = - \sum_{k=1}^{K} S_{k}
\label{supp:eq:discrete_survival_model_risk_score}
\end{align}
where $S_{k}$ represents the predicted probability that the patient survives interval $k$. This survival head structure is illustrated in Figure~\ref{fig:mil-model-heads}-b. This formulation allows for training a model to assign survival interval risks per individual patient instead of comparing samples at training time (as required in the Cox partial log-likelihood). The discrete-time survival model is commonly optimized via its log-likelihood \citep{chen21mcat, zadeh2020bias}, which does not require comparing multiple samples and hence also works for small or single-sample mini-batches.
\subsection{Survival cost function}\label{supp:sec:survival_cost}

Assume our observed data for the $i$-th patient is in the form of $(X_i, c_i, y_i)$ where $X_i$ is the bag of instances, $c_i$ is the censorship status ($c_i=1$ means censored data), and $y_i \in \{1, \ldots, K\}$ is the discretized event time as described in the previous section. Additionally, we denote the random variable for the event time as $T_i$. For the sake of simplifying the notation, we drop the index $i$.

The probability of experiencing the event in interval $k$ given survival up to $k$ is $f_{\text{hazard}}(k \mid X) = P(T = k \mid T \ge k,\, X)$. The probability of surviving beyond interval k is then:
\begin{equation}
    f_{\text{survival}}(k \mid X) = P(T > k \mid X) = \prod_{u=1}^{k} \big[ 1 - f_{\text{hazard}}(u \mid X) \big]
\end{equation}

Then, the log-likelihood for this patient is \citep{zadeh2020bias}:
\begin{equation}
    L = -\, c \cdot \log f_{\text{survival}}(y) - (1 - c)\big[\, \log f_{\text{survival}}(y - 1) + \log f_{\text{hazard}}(y) \big]
\end{equation}

To increase the influence of uncensored cases, we add a weighting factor \citep{chen21mcat}:
\begin{equation}
\begin{gathered}
     L_{\text{total}} = (1 - \beta) L + \beta L_{\text{uncensored}} \\
    L_{\text{uncensored}} = - (1 - c)\big[\, \log f_{\text{survival}}(y - 1) + \log f_{\text{hazard}}(y) \big]
\end{gathered}
\end{equation}

This formulation ensures that both censored and uncensored patients contribute appropriately to the loss while placing greater emphasis on fully observed survival times.


\section{Explanation methods} \label{supp:sec:exp_methods}
\subsection{Details of the methods}
\paragraph{Attention heatmaps} 
In Attention MIL (AttnMIL), the attention score assigned to each instance represents its contribution to the bag-level prediction. In Transformer-based MIL models, self-attention matrices are computed at each layer, and the attention scores from the class token to the instance tokens are extracted. To aggregate information across layers, attention rollout \citep{abnar-zuidema-2020-quantifying} is often used, which recursively combines the attention matrices. 

Formally, given that the attention heads deliver self-attention vectors $\mathbf{A}_{h}^{l} \in \mathbb{R}^{(K + 1) \times (K + 1)}$ for each head $h$ and Transformer layer $l$, recalling that the first token is the class token. Mean pooling is often used for fusing the self-attention matrices of different heads, i.e., $\mathbf{A}^l = \left \langle \mathbf{A}_h^l \right \rangle_h$. The attention scores from the class token to the instance tokens can be used as attribution scores, i.e., $\mathbf{A}^l_{(1, 2:)}$. Alternatively, for a model with $L$ Transformer layers, attention rollout combines $\{\mathbf{A}^l\}_{l=1}^{L}$ as $\tilde{\mathbf{A}}=\prod_{l=1}^{L} \check{\mathbf{A}}^l$ where $\check{\mathbf{A}}^l=0.5\mathbf{A}^l + 0.5\mathbf{I}$, with $\mathbf{I}$ being the identity matrix. Then, similar to the layer-wise attention scores, the heatmap is defined as the attention rollout of the class token to the instances, i.e.,  $\tilde{\mathbf{A}}_{(1, 2:)}$.

The direct interpretability of attention scores is insufficient to faithfully reflect the model predictions \citep{ali2022xai, jain-wallace-2019-attention, wiegreffe-pinter-2019-attention}. Moreover, they cannot distinguish between positive, negative, or class-wise evidence \citep{hense2024xmil, javed2022additivemil}.

\paragraph{Gradient-based methods}
While not frequent, Integrated Gradients (IG) is the most used gradient-based method in the histopathology MIL literature \citep{lee2022derivation, li2023vision, song2024analysis, vu2025contrastiveig}. IG \citep{sundararajan2017axiomatic} computes the gradients of the model's output with respect to the input, integrated over a path from a baseline to the actual input. The baseline is typically set to zero, and so we do. The explanation score of the $k$-th instance is computed as $\sum_d \text{IG}(x_{kd})$, where the relevance score of the $d$-th feature of the $k$-th instance $\text{IG}(x_{kd})$ is computed as  
\begin{equation}
    \text{IG}(x_{kd}) = x_{kd} \cdot \int_{\alpha=0}^{1} \frac{f( \alpha \mathbf{X})}{\partial x_{kd}} d\alpha,
\end{equation}
where $f$ is the model and $\mathbf{X}$ is the $K\times D$ feature matrix of the bag with the instance feature vectors on its rows.  We used the implementation of IG available in Captum \citep{kokhlikyan2020captum} with the internal batch size set to the number of instances in a bag.

Gradient$\,\times\,$Input (G$\times$I) \citep{baehrens10a,DBLP:journals/corr/ShrikumarGSK16} is a common baseline method in XAI literature. The G$\times$I heatmap score of the $k$-th instance is defined as $h_k = \nabla_{\mathbf{x}_k} f(X)^T \mathbf{x}_k$, where $\nabla_{\mathbf{x}_k} f(X) \in \mathbb{R}^D$ is the gradient of the model output with respect to the $k$-th instance features $\mathbf{x}_k \in \mathbb{R}^D$.

\paragraph{Perturbation-based methods}
Perturbation-based methods, building on model-agnostic approaches like SHAP \citep{lundberg2017shap}, perturb bag instances and compute importance scores from the resulting change in the model prediction. \cite{early2022shapmil} proposed and evaluated multiple perturbation-based methods of different complexity. The ``Single'' method passes bags of single patches $X_k = \{ \mathbf{x}_{k} \}$ for $k=1,\ldots,K$ through the model and uses the outcome $f(\mathbf{x}_k)$ as explanation score. ``One removed'' drops single patches, i.e., constructs bags $\check{X}_{k} = X \backslash \mathbf{x}_k$ for $k=1,\ldots,K$ and defines the difference to the original prediction score $f(X) - f(\check{X}_k)$ as explanation. The ``combined'' approach takes the mean of these two scores. As these methods cannot account for patch interactions, \cite{early2022shapmil} also propose an algorithm to sample coalitions of patches to be perturbed, called MILLI.

Some of these approaches come with high computational complexity. The linear ``Single'' and ``one removed'' methods require $K$ forward passes per bag. While ``Single'' only passes small bags through the model, which speeds up computation, ``one removed'' requires $K$ passes with almost full bag size. MILLI scales quadratically with the number of instances \citep{early2022shapmil}. In histopathology, where bags typically contain more than 1,000 and frequently more than 10,000 instances, quadratic runtime is practically infeasible. 

\paragraph{Layerwise Relevance Propagation (LRP)}
LRP \citep{bach-plos15, lapuschkin2019unmasking, montavon2018methods, montavon2019layer, samek2021-xai} explains neural network predictions by redistributing the output's relevance back through the network to the input features. The redistribution follows a relevance conservation principle, where the total relevance of each layer is preserved as it propagates backward. If $r_j^{(l)}$ denotes the relevance of neuron $j$ in layer $l$, conservation means that $\sum_{j} r_j^{(l_1)} = \sum_{i} r_i^{(l_2)}$ holds for any two layers $l_1$ and $l_2$. As a general principle, LRP posits
\begin{equation}
    r_i^{(l)} = \sum_j \frac{q_{ij}}{\sum_{i'} q_{i'j}}\cdot r_j^{(l+1)},
\end{equation}
with $q_{ij}$ being the contribution of neuron $i$ of layer $l$ relevance $r_j^{(l+1)}$. There are ``propagation rules'' for various layer types \citep{montavon2018methods, montavon2019layer} that specify $q_{ij}$ for different setups. For the attention mechanism, as a core component of many MIL architectures, we employ the AH-rule introduced by \cite{ali2022xai}. In a general attention mechanism, let $\mathbf{z}_k=\left[z_{kd}\right]_d$ be the embedding vector of the $k$-th token and $p_{kj}$ the attention score between tokens $k$ and $j$. The output vector of the attention module is $\mathbf{y}_j = \sum_{k} p_{kj} \mathbf{z}_k$. The AH-rule of LRP treats attention scores as a constant weighting matrix during the backpropagation pass of LRP. If $R(y_{jd})$ is the relevance of the $d$-th dimension of $\mathbf{y}_j=\left[y_{jd}\right]_d$, the AH-rule computes the relevance of the $d$-th feature of $\mathbf{z}_k$ as: 
\begin{equation}
    R(z_{kd})=\sum_{j} \frac{z_{kd} p_{kj}}{\sum_{i} z_{id}p_{ij}} R(y_{jd}).
\end{equation}

This formulation can be directly applied to AttnMIL, and also adapted to a QKV attention block in a transformer, where $\mathbf{z}_k$ is the embedding associated with the value representation.

For Mamba layers, we follow \cite{NEURIPS2024mambaLRP}. In Mamba models, the relevance conservation property is violated in three components: the SiLU activation function, the selective SSM, and the multiplicative gate. To restore the conservation property in the relevance propagation pass, we perform a locally linear expansion of these components by treating certain quantities as constants. For the SiLU activation function, we treat the sigmoid factor as constant, for the selective SSM, we treat the parameter matrices as constants and for the multiplicative gate, we treat half of the factors as constant. Under the expansion, the components can be treated as linear layers and relevance propagation for linear layers can be applied.

We further implement the LRP-$\epsilon$ rule for linear layers followed by ReLU activation function \citep{montavon2018methods}, as well as the LN-rule to address the break of conservation in layer norm \citep{ali2022xai}.

At the instance-level, LRP assigns each instance $\mathbf{x}_k=[x_{kd}]_d \in \mathbb{R}^D$ a relevance vector $\mathbf{r}_k=[r_{kd}]_d$ with $r_{kd}=R(x_{kd})=r^{(0)}_{kd}$ being the relevance score of the $d$-th feature of $\mathbf{x}_k$. We define the instance-wise relevance score as an estimate for the evidence score of the instance as $h_k = \sum_d r_{kd}$.

\subsection{LRP rules} \label{supp:sec:lrp_rules}

\textbf{Feed forward neural network}. The following generic rule holds for propagating relevance through linear layers followed by ReLU \cite{montavon2019layer}:
\begin{equation} \label{supp:eq:linear_lrp_rule}
    r_i^{(l)} = \sum_j \frac{a_j \rho(w_{ij})}{\epsilon + \sum_{i'} a_{i'} \rho(w_{i'j})}\cdot r_j^{(l+1)},
\end{equation}
where $a_{j}$ is the activation of neuron $j$ in layer $l$, $w_{ij}$ the weight from neuron $i$ of layer $l$ to neuron $j$ of layer $l+1$, $\epsilon$ a stabilizing term to prevent numerical instabilities, and $\rho(w_{ij})$ a modification of the weights of the linear layer. For example, if $\rho(w_{ij})=w_{ij} + \gamma \text{max}(w_{ij}, 0)$, then Equation \ref{supp:eq:linear_lrp_rule} is called LRP-$\gamma$ rule. For $\gamma=0$, this equation is called LRP-$\epsilon$ rule.

\textbf{LayerNorm}. Assume $\mathbf{z}_k$ is the embedding of the $k$-th token and $\mathbf{y}_k=\text{LayerNorm}(\mathbf{z}_k)$ as:
\begin{equation}
    \mathbf{y}_k=\frac{\mathbf{z}_k - \text{E}\{\mathbf{z}\}}{\text{std}\{\mathbf{z}\} + \epsilon},
\end{equation}
where $\text{E}\{\mathbf{z}\}$ and $\text{std}\{\mathbf{z}\}$ are the expected values and standard deviation of the tokens. 

For propagating relevance through LayerNorm, \cite{ali2022xai} suggested the LN-rule as the following:
\begin{equation}
     R(z_{kd}) = \sum_{j} \frac{z_{kd} (\delta_{kj} - \frac{1}{N})}{\sum_{i} z_{id} (\delta_{ij} - \frac{1}{N})} R(y_{jd}),
\end{equation}
where $\delta_{kj} = \begin{cases} 1, & \text{if } k = j \\ 0, & \text{otherwise} \end{cases}$ and  $z_{kd}$ is the $d$-th dimension of $\mathbf{z}_k$ and $R(z_{kd})$ is the relevance assigned to it. In practice, LN-rule is implemented by detaching $\text{std}\{\mathbf{z}\}$ and handling it as a constant.

\textbf{SiLU activation function}. 
The SiLU activation function $y_i = x_i\ \cdot \sigma(x_i)$ is locally linear expanded by treating $\sigma(x_i)$ as constant. The resulting rule for the SiLU activation function is
\begin{equation}
    \mathcal{R}(x_i) = \mathcal{R}(y_i)
\end{equation}

\textbf{Selective SSM}.
The selective SSM 
\begin{align}
    h^{(t)} &= A^{(t)}h^{(t-1)} + B^{(t)}x^{(t)} \\
    y^{(t)} &= C^{(t)}h^{(t-1)}
\end{align}
is locally linear expanded by treating $A$, $B$ and $C$ as constants. The resulting rules for the selective SSM are
\begin{align}
    \begin{split}
        \mathcal{R}\left(h_i^{(t-1)}\right) &= \sum_j \frac{h_i^{(t-1)}\left[C^{(t-1)}\left(h^{(t-1)}\right)\right]_{ji}}{\sum_{i'} h_{i'}^{(t-1)}\left[C^{(t-1)}\left(h^{(t-1)}\right)\right]_{ji'}} \mathcal{R}\left(y_j^{(t-1)}\right) \\ &+ \sum_j \frac{h_i^{(t-1)}\left[A^{(t)}\left(x^{(t)}\right)\right]_{ji}}{\sum_{i'} x_{i'}^{(t)}\left[A^{(t)}\left(x^{(t)}\right)\right]_{ji'} + \sum_{i''} x_{i''}^{(t)}\left[B^{(t)}\left(x^{(t)}\right)\right]_{ji''}} \mathcal{R}\left(h_j^{(t)}\right)
    \end{split} \\
    \mathcal{R}\left(x_i^{(t)}\right) &= \sum_j \frac{x_i^{(t)} \left[B^{(t)}\left(x^{(t)}\right)\right]_{ji}}{\sum_{i'}x_{i'}^{(t)}\left[B^{(t)}\left(x^{(t)}\right)\right]_{ji'} + \sum_{i''} h_{i''}^{(t-1)}\left[A^{(t)}\left(x^{(t)}\right)\right]_{ji''}}\mathcal{R}\left(h_j^{(t)}\right)
\end{align}

\textbf{Multiplicative Gate}. The multiplicative gate $y = z_A \cdot z_B$ is locally linear expanded by treating half of $z_a \cdot z_B$ as constant. The resulting rules for the multiplicative gate is
\begin{equation}
    \mathcal{R}([z_A]_i) = 0.5 \cdot \mathcal{R}(y_i)
\end{equation}
\begin{equation}
    \mathcal{R}([z_B]_i) = 0.5 \cdot \mathcal{R}(y_i)
\end{equation}

\section{Details of the large-scale evaluation experiments}\label{supp:sec:experimental_setup}

\subsection{Datasets and prediction tasks}\label{supp:sec:experimental_setup:datasets}

We consider all three common machine learning tasks in digital histopathology, namely classification, regression, and survival prediction.
\paragraph{Classification}
For the classification task, we include datasets that cover tumor detection, disease subtyping, and biomarker prediction. 

\begin{itemize}
\item 
The first task is a tumor detection task using the CAMELYON16 \citep{bejnordi2017camelyon}, which consists of 400 sentinel lymph node slides, of which 160 carry to-be-recognized metastatic lesions of different sizes. The goal is to distinguish tumor-containing slides from normal tissue.
\item 
The second task focuses on disease subtyping using TCGA NSCLC (abbreviated as NSCLC), which includes 529 slides with lung adenocarcinoma (LUAD) and 512 with lung squamous cell carcinoma (LUSC). The objective is to classify these two non-small cell lung cancer (NSCLC) subtypes.
\item 
The third task involves biomarker prediction using TCGA HNSC. This dataset comprises 433 slides of head and neck squamous cell carcinoma (HNSC), 43 of which are positive for human papillomavirus (HPV) infection, confirmed via additional testing \citep{campbell2018tcgascc}. HPV infection is an important biomarker for prognosis and treatment planning \citep{bilal2023aggregation}. The task is to predict HPV status directly from the H\&E slides. This challenging task, referred to as HNSC HPV throughout the paper, is characterized by label imbalances and a subtle predictive signature.
\item 
The fourth task is another biomarker prediction, this time using TCGA LUAD. This dataset contains 529 lung adenocarcinoma (LUAD) slides, 263 of which show a TP53 mutation--one of the most common mutations in cancer. In lung cancer, TP53 mutations are associated with poorer prognosis and resistance to therapy \citep{mogi2011tp53}. Prior studies have shown that TP53 status is predictable from LUAD histology slides \citep{coudray2018nsclc, wang2021heal}. The task is abbreviated as LUAD TP53 hereafter.
\end{itemize}
\paragraph{Regression}
For the regression task, we used the TCGA BRCA dataset and trained MIL regression models to predict two continuous targets, including gene expression as well as biomarker prediction.
\begin{itemize}
\item 
In the first task, the model was trained to predict the expression level of AURKA, measured in fragments per kilobase million with upper-quartile normalization (FPKM-UQ), obtained from the UCSC Xena Browser \citep{goldman2020visualizing}. A total of 1,081 patients had the gene expression value available. Prior work of \cite{wang2021predicting} shows that AURKA expression can be inferred from H\&E slides, achieving a Spearman correlation of 0.66 on the TCGA BRCA internal test set. Here, we used the raw FPKM-UQ values to train a regression model. This approach implicitly assumes a reference value of zero, corresponding to the absence of detectable gene expression.

\item 
The second task involved predicting homologous recombination deficiency (HRD) scores from H\&E slides of TCGA BRCA cohort. HRD is considered a predictive pan-cancer biomarker that may help identify patients who could benefit from PARP inhibitor (PARPi) therapy. However, HRD testing remains technically complex despite its clinical importance \citep{loeffler2024prediction}. Prediction of HRD scores using Attention MIL regression models has been shown to be feasible \citep{el2024regression}, reporting a Pearson's correlation of 0.53 in this dataset. We obtained the HRD values for the TCGA BRCA cohort from the supplementary data provided by \cite{loeffler2024prediction}. A total of 1,026 cases had available HRD measurements. A clinical reference value of 42 is used for classifying a test to HRD+ and HRD-. We used this reference value to predict the difference of measured HRD to the reference value. This way, we adapt the methodology in \citep{letzgus2022} for explaining regression models. The heatmaps of such a model would show how the patches contribute to the sample to move towards HRD+ and HRD-. 

\end{itemize}
\paragraph{Survival}
Four different datasets from TCGA have been used for predicting overall survival, i.e., for cancers of the breast (BRCA), the kidney (KIRC), the brain (GBM \& LGG), and the colorectal region (COAD \& READ). Survival data were available for at least 537 (KIRC) and up to 1,096 (BRCA) patients. Overall survival has frequently been predicted on these datasets, and they have served as a benchmark for (multimodal) aggregation models (see, e.g., \cite{chen21mcat, chen2022pan, song24mmp, eijpe2025dimaf}), in which unimodal histopathology MIL models frequently achieved C-Index scores between $0.55$ and $0.8$.

\subsection{H\&E slides preprocessing}

We extracted patches from the slides of $256 \times 256$ pixels without overlap at 20x magnification (0.5 microns per pixel) using OpenSlide\footnote{\url{https://openslide.org/}}. We identified and excluded background patches via Otsu's method \citep{otsu1975threshold} on slide thumbnails and applied a patch-level minimum standard deviation of 8.

\subsection{Foundation models}

To investigate whether using different feature extractors impacts MIL explanations, we included two state-of-the-art models, namely UNI2-h\footnote{\url{https://huggingface.co/MahmoodLab/UNI2-h}} \citep{chen2024uni} and Virchow2 \citep{zimmermann2024virchow2} in our experiments. Both models were pre-trained on a massive number of slides (UNI2-h: 350K slides, Virchow2: 3.1M slides) and consistently ranked among the best-performing models in various patch-level and slide-level prediction tasks \citep{campanella2025benchmark, kaiko2024eva, jaume2024hest, neidlinger2025benchmarking}. Virchow2 was also found to be one of the foundation models most robust to technical variations (e.g., staining and scanner differences) \cite{koemen2025pathorob}, which can be crucial to obtain biologically meaningful heatmaps from multi-hospital training data.

\subsection{Model training/testing and model selection}

We used a 5-fold cross-validation scheme with patient-level splitting, with a hyperparameter grid of the models trained in this work are reported in \ref{supp:sec:experimental_setup:hyperparams}. 

In this cross-validation scheme, the dataset was split into five partitions. For each of the five iterations, one partition was used as the test set, the next (cyclically) as the validation set, and the remaining three as the training set. Each fold was therefore used exactly once for validation and once for testing, which means that validation and test sets overlap across iterations. Therefore, to avoid data leakage, hyperparameter selection was performed independently within each iteration, using only that iteration’s validation set — resulting in potentially different hyperparameters for each fold.

For training, we sampled bags of 2,048 patches per slide and passed their feature representations through the MIL model. For validation and testing, we simultaneously exposed all patches of a slide to the model to avoid sampling biases and statistical instabilities. The checkpoint with the best performance metric on the validation set was saved during the training process. 

\paragraph{Task-specific cost functions}
For classification tasks, we use the standard cross-entropy loss between the predicted class logits and the ground-truth label. 
For regression tasks, the mean squared error is used to measure the difference between the predicted and true continuous target values.
For survival prediction, we use a discrete-time log-likelihood loss as detailed in \ref{supp:sec:survival_cost}.

\paragraph{Performance metrics} 
For classification, regression, and survival prediction models, we used the area under the receiver operating characteristic curve (AUROC), Spearman correlation, and C-Index as the performance metrics, respectively.

\subsection{Hyperparameter grids}\label{supp:sec:experimental_setup:hyperparams}
All the hyperparameter grids for all the models trained in this work are presented in Table \ref{supp:tab:hyperparam-grids}.

\begin{table}[H]  
\centering
\caption{Hyperparameter grids for all models trained in this work}
\label{supp:tab:hyperparam-grids}
\begin{tabularx}{\textwidth}{llX}
\toprule
\textbf{Task} & \textbf{Model} & \textbf{Hyperparameter Grid} \\
\midrule

\multirow{3}{*}{Classification}
  & TransMIL & train bag size=\{2048\}, batch size=\{5\},  lr=\{2e-4, 2e-3\}, wd=\{0, 0.01\}, dropouts=\{(0, 0, 0), (0.5, 0.5, 0.2)\}, opt=\{SGD\}, epochs=200, warmup=0 \\
  
  & AttnMIL & train bag size=2048, batch size=32,  lr=\{0.002, 0.0002\}, wd=\{0, 0.01\}, dropouts=\{0, 0.5\}, opt=SGD, epochs=1000, warmup=0 \\
  
  & MambaMIL & train bag size=\{2048\}, batch size=\{5\},  lr=\{2e-4, 2e-3\}, wd=\{0,0.01\},opt=\{SGD\}, epochs=200, warmup=0 \\

\midrule

\multirow{3}{*}{Regression}
  & TransMIL & train bag size=\{2048\}, batch size=\{8\},  lr=\{1e-6, 1e-5, 1e-4\}, wd=\{0.01, 0.1\}, dropouts=\{(0.2, 0.5, 0.2), (0.5, 0.75, 0.5)\}, opt=\{SGD, Adam\}, epochs=300, warmup=5 \\
  
  & AttnMIL &  train bag size=\{2048\}, batch size=\{32\},  lr=\{2e-5, 2e-4\}, wd=\{0.01, 0.1\}, dropouts=\{0.2, 0.75\}, opt=\{SGD, Adam\}, epochs=150, warmup=5 \\
  
  & MambaMIL & train bag size=\{2048\}, batch size=\{5\},  lr=\{1e-6, 1e-5, 1e-4\}, wd=\{0, 0.01\}, opt=\{SGD\}, epochs=200, warmup=0 \\

\midrule

\multirow{3}{*}{Survival}
 & TransMIL & train bag size=2048, batch size=5,  lr=\{0.002, 0.0002\}, wd=\{0, 0.01\}, dropouts=\{(0, 0, 0), (0.5, 0.5, 0.2)\}, opt=SGD, epochs=200, warmup=0 \\
  
  & AttnMIL & train bag size=2048, batch size=32,  lr=\{0.002, 0.0002\}, wd=\{0, 0.01\}, dropouts=\{0, 0.5\}, opt=SGD, epochs=1000, warmup=0 \\
  
  & MambaMIL & train bag size=\{2048\}, batch size=\{5\},  lr=\{2e-4, 2e-3\}, wd=\{0, 0.01\}, opt=\{SGD\}, epochs=200, warmup=0 \\

\midrule

\multirow{1}{*}{HEST-Regression}
& TransMIL & batch size=\{32\},  lr=\{0.001, 0.0001, 0.00001\}, wd=\{0.001, 0.01, 0.1\}, dropouts=\{(0, 0.25, 0.5, 0.75), (0, 0.25, 0.5, 0.75), (0, 0.2, 0.4, 0.6)\}, opt=\{SGD\} \\
 
\bottomrule

\end{tabularx}
\end{table}

\subsection{Model performance tables}

\setlength{\tabcolsep}{2pt}
\begin{table}[H]
\scriptsize
\centering
\caption{Performance results for the models in faithfulness experiments, with backbone Virchow2. The numbers are mean (std) of the performance over the cross-validation folds.}
\label{supp:tab:model_performances_virchow}
\begin{tabular}{lcccc|cc|cccc}
\toprule
{} & \multicolumn{4}{c}{Classification} & \multicolumn{2}{c}{Regression } &\multicolumn{4}{c}{Survival} \\
\cmidrule(lr){2-5} \cmidrule(lr){6-7} \cmidrule(lr){8-11}
 & CAMELYON & NSCLC & HPV & TP53 & AURKA & HRD & BRCA & GBMLGG & KIRC & COADREAD \\
\midrule
TransMIL & 1.00(0.00) & 0.97(0.02) & 0.90(0.04) & 0.71(0.05) & 0.66(0.02) & 0.62(0.02) & 0.65(0.08) & 0.74(0.04) & 0.70(0.03) & 0.63(0.06) \\
AttnMIL & 1.00(0.00) & 0.98(0.01) & 0.92(0.04) & 0.68(0.07) & 0.69(0.03) & 0.60(0.02) & 0.62(0.05) & 0.76(0.04) & 0.71(0.03) & 0.61(0.07) \\
MambaMIL & 0.97(0.03) & 0.90(0.04) & 0.89(0.03) & 0.63(0.08) & 0.65(0.03) & 0.60(0.03) & 0.59(0.05) & 0.76(0.12) & 0.68(0.06) & 0.64(0.06) \\
\bottomrule
\end{tabular}
\end{table}
\setlength{\tabcolsep}{2pt}
\begin{table}[H]
\scriptsize
\centering
\caption{Performance results for the models in faithfulness experiments, with backbone UNI2. The numbers are mean (std) of the performance over the cross-validation folds.}
\label{supp:tab:model_performances_uni}
\begin{tabular}{lcccc|cc|cccc}
\toprule
{} & \multicolumn{4}{c}{Classification} & \multicolumn{2}{c}{Regression } &\multicolumn{4}{c}{Survival} \\
\cmidrule(lr){2-5} \cmidrule(lr){6-7} \cmidrule(lr){8-11}
 & CAMELYON & NSCLC & HPV & TP53 & AURKA & HRD & BRCA & GBMLGG & KIRC & COADREAD \\
\midrule
TransMIL & 1.00(0.00) & 0.98(0.01) & 0.92(0.04) & 0.68(0.08) & 0.66(0.03) & 0.62(0.05) & 0.65(0.07) & 0.75(0.04) & 0.71(0.06) & 0.62(0.08) \\
AttnMIL & 1.00(0.00) & 0.97(0.02) & 0.92(0.04) & 0.65(0.07) & 0.69(0.03) & 0.63(0.03) & 0.58(0.05) & 0.75(0.03) & 0.71(0.04) & 0.58(0.06) \\
MambaMIL & 0.98(0.02) & 0.86(0.03) & 0.88(0.04) & 0.62(0.07) & 0.66(0.04) & 0.61(0.04) & 0.62(0.06) & 0.75(0.12) & 0.71(0.07) & 0.62(0.09) \\
\bottomrule
\end{tabular}
\end{table}

\section{Details of statistical analyses}\label{supp:sec:stat_details}

We assess the pairwise differences of explanation methods using Wilcoxon's signed-rank test. We compute the effect size using the following equation \citep{RAMACHANDRAN2021491}:
\begin{align}
    r = \frac{z}{\sqrt{n}} \\
    z = \frac{W^+ - \mu}{\sigma} \\
    \mu = \frac{n(n+1)}{4} \\
    \sigma = \sqrt{n(n+1)(2n+1)}{24}
\end{align}
where $n$ is the number of non-zero paired differences, $W^+$ is the sum of positive ranks of the paired differences.

\begin{figure}[t!]
    \centering
    \includegraphics[width=\linewidth]{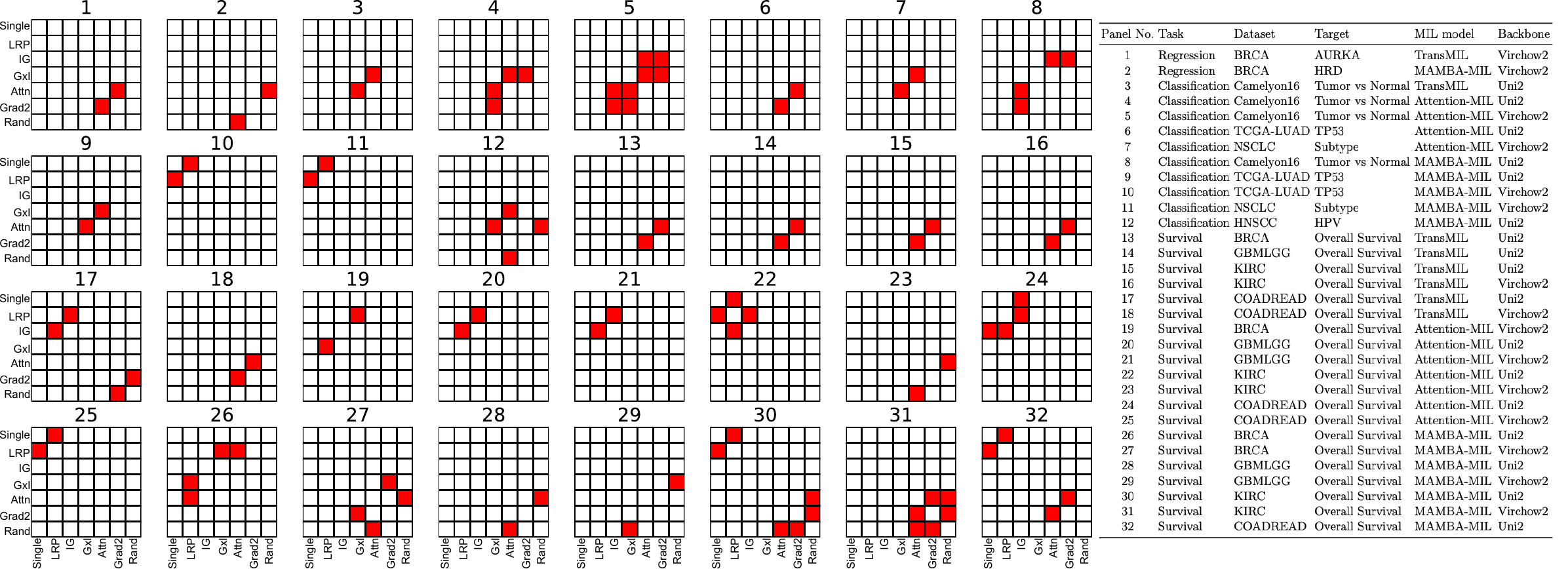}
    \caption{\textbf{Experimental configurations with at least one non-significant pairwise faithfulness differences after multiple comparison correction.} Each panel corresponds to one experimental configuration (dataset × task × backbone × MIL architecture) in which at least one pairwise Wilcoxon signed-rank test between explanation methods was non-significant after FDR correction ($\alpha = 0.05$). Cells indicate method pairs, and the red color marks the pairs for which no statistically significant difference in SRG faithfulness was detected. Most non-significant comparisons occur within the two groups \{Single, LRP, IG\} and \{Attention, G×I, Grad2, Random\}. Only two configurations in panels 4 and 7 (both CAMELYON16 tumor detection experiments) exhibit non-significant differences across groups, consistent with the simplicity of the task and the aggregation-level results reported in Section~\ref{sec:results:faithfulness_stat_comparison}.}
    \label{supp:fig:non-significant-pairwise-p-values}
\end{figure}

We performed 21 pairwise statistical tests for each experimental configuration (corresponding to all method pairs), resulting in 1,260 tests in total across all configurations. After FDR correction for controlling for multiple comparisons, we used 0.05 as the significance level. Figure~\ref{supp:fig:non-significant-pairwise-p-values} depicts all individual experimental configurations in which at least one non-significant pairwise Wilcoxon signed-rank test existed after FDR correction. Nearly all non-significant cases occur within either of the groups \{Single, LRP, IG\} and \{Attention, G×I, Grad2, Random\}. Two configurations deviate from this pattern, both corresponding to CAMELYON16 tumor detection experiments: panel 4 is for AttnMIL, and panel 7 is for MambaMIL. In the AttnMIL panel (4), \{IG, Attn, Grad2, GI\} show no significant differences, whereas in the MambaMIL panel (7), \{IG, Attn, Grad2\} are not significantly different in their faithfulness. These observations are consistent with the inferential results reported in Section~\ref{sec:results:faithfulness_stat_comparison} and Figure~\ref{fig:faithfulness_aggregated}, which indicate reduced separability between explanation methods for simpler tasks and architectures.

\begin{figure}[t!]
    \centering
    \includegraphics[width=\linewidth]{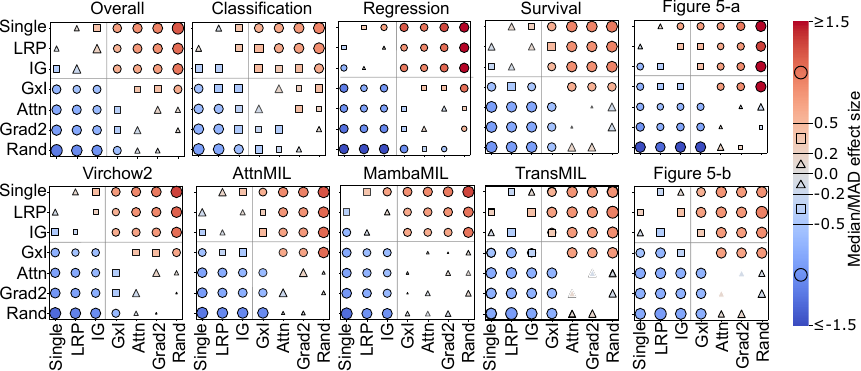}
    \caption{\textbf{Pairwise median/MAD-based comparison of SRG scores between explanation methods, averaged over various experimental factors.} This figure corresponds to Figure~\ref{fig:faithfulness_aggregated}.  The filling colors indicate the magnitude of median/MAD as shown on the colorbar. Triangle, square, and circle markers are used to denote the negligible ($< 0.2$), weak to moderate ($0.2-0.5$), and moderate to strong ($\geq 0.5$) effect sizes.}
    \label{supp:fig:faithfulness_aggregated_median}
\end{figure}

To assess the robustness of the pairwise faithfulness comparisons to the choice of effect size metric, we replicated all aggregated analyses discussed in Section~\ref{sec:results:faithfulness_stat_comparison} and Figure~\ref{fig:faithfulness_aggregated} using an alternative standardized measure of paired differences (Figure~\ref{supp:fig:faithfulness_aggregated_median}), namely the standardized median defined as $\mathrm{median}(d)/(1.4826 \cdot \mathrm{MAD}(d))$, with $d$ being the paired differences. We abbreviate it as median/MAD. In addition, we reproduced the center panels of Figure 5a–b using the median/MAD effect size instead of Wilcoxon’s signed-rank effect size. 

Across all aggregation levels (task type, MIL architecture, backbone, and overall), the median/MAD results closely match the patterns observed with Wilcoxon’s signed-rank effect size. In particular, the consistent grouping of explanation methods \{Single, LRP, IG\} versus \{Attention, G×I, Grad2, Random\} and the relative ordering within these groups remained stable. 

Overall, these supplementary analyses demonstrate that the reported faithfulness comparisons are robust to the choice of effect size definition and do not depend on a specific scale normalization.

\section{Details for spatial transcriptomics validation experiment}
\label{supp:sec:HEST_training_details}

HEST-1k is a publicly available dataset comprising 1,229 paired H\&E WSI and spatial transcriptomics (ST) samples derived from two species \citep{jaume2024hest}. For each pair, the ST data were spatially registered to the corresponding H\&E slide. In this study, experiments were conducted on the subset defined as the \textit{breast cancer task} in the original publication, which consists of four H\&E–ST pairs obtained from four independent patients. All slides were formalin-fixed, paraffin-embedded (FFPE) tissue sections profiled using the Xenium pipeline (v1).

To ensure patient-level separation and avoid data leakage, a four-fold cross-validation scheme was implemented, with each fold corresponding to one patient (i.e., one slide). We trained a TransMIL-based regression framework using a Virchow2 backbone using the pipelines described in other sections. Hyperparameters were optimized via grid search (see Table \ref{supp:sec:HEST_training_details}). For downstream visualization, heatmaps were generated from the checkpoint achieving the lowest validation loss within each fold.

\begin{figure}[H]
  \centering
  \includegraphics[width=.8\linewidth]{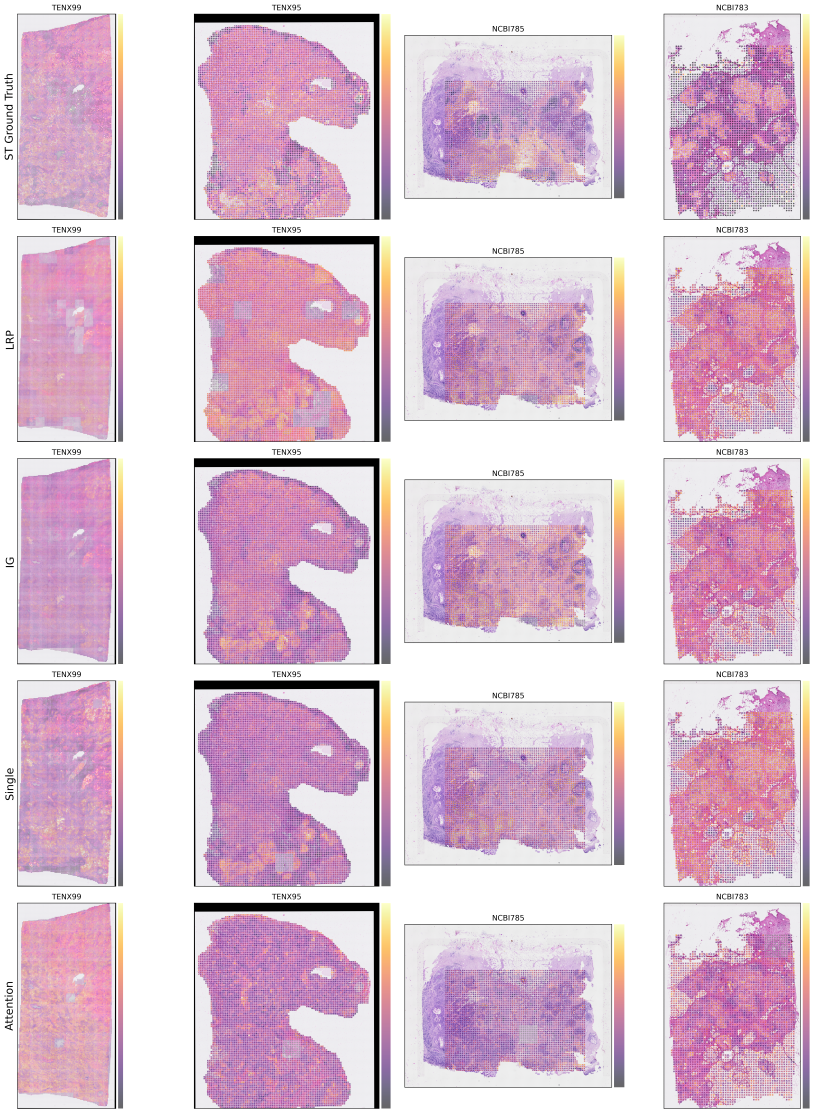}
\caption{\textbf{Ground truth and heatmaps for the ERBB2 prediction model in Section~\ref{sec:hest_biological_val}}. The ground-truth spatial transcriptomics and the heatmaps for the four explanation methods used in the ST validation study presented in Section~\ref{sec:hest_biological_val}.}
\label{supp:fig:HEST_all_heatmaps_ERBB2}
\end{figure}

\begin{figure}[H]
  \centering
  \includegraphics[width=.8\linewidth]{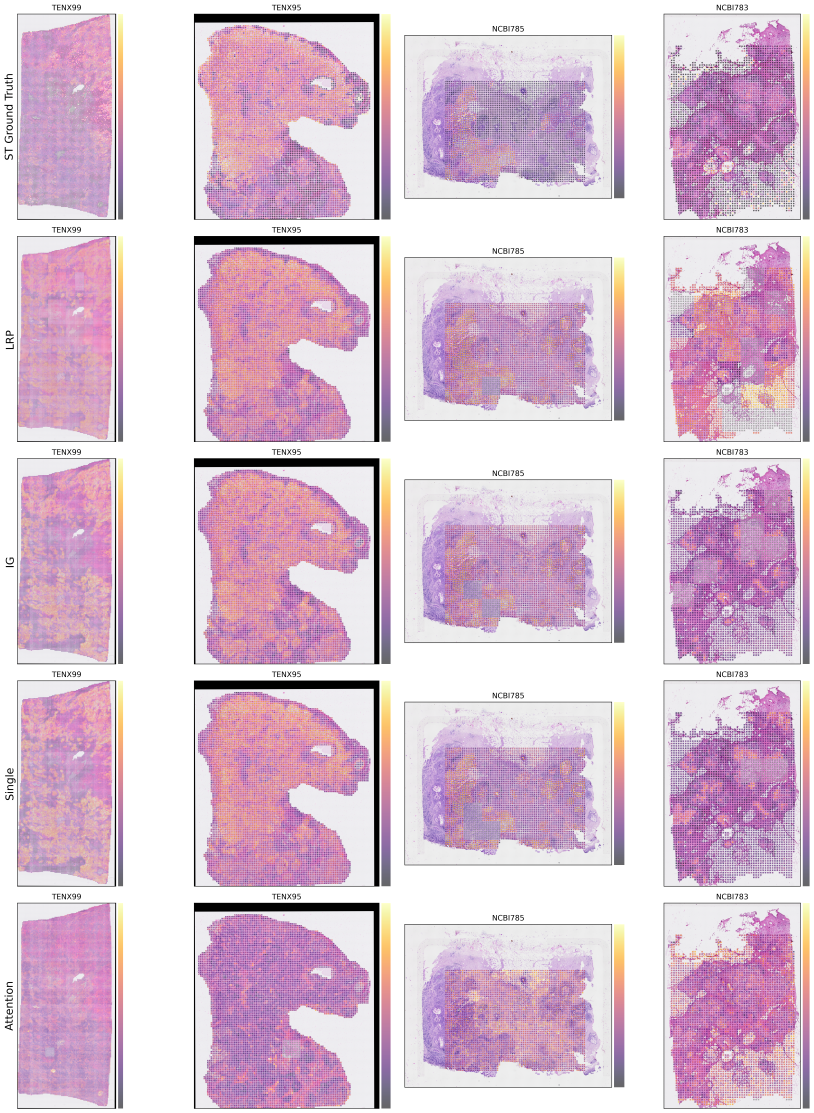}
\caption{\textbf{Ground truth and heatmaps for the FASN prediction model in Section~\ref{sec:hest_biological_val}}. The ground-truth spatial transcriptomics and the heatmaps for the four explanation methods used in the ST validation study presented in Section~\ref{sec:hest_biological_val}.}
\label{supp:fig:HEST_all_heatmaps_FASN}
\end{figure}

\section{Details for head and neck cancer discovery analysis}
\label{sec:supp:use_case_2_details}

For tissue compartment segmentation, we use the same approach as in \cite{hense2025pathways}. Specifically, we apply a patch classification model distinguishing between tumor and non-tumor patches. The model is a lightweight classifier on top of RudolfV foundation model features. It was trained on pixel-level tumor region annotations for 60 slides and validated with 15 slides from a proprietary head and neck cancer cohort from LMU Munich (see \cite{hense2025pathways} for details). To increase spatial consistency, a neighbourhood smoothing algorithm is applied to the patch prediction scores. Border patches were determined via a specified border width of two patches distance from any predicted tumor patch.

\begin{figure}[!t]
    \centering
    \includegraphics[width=0.95\linewidth]{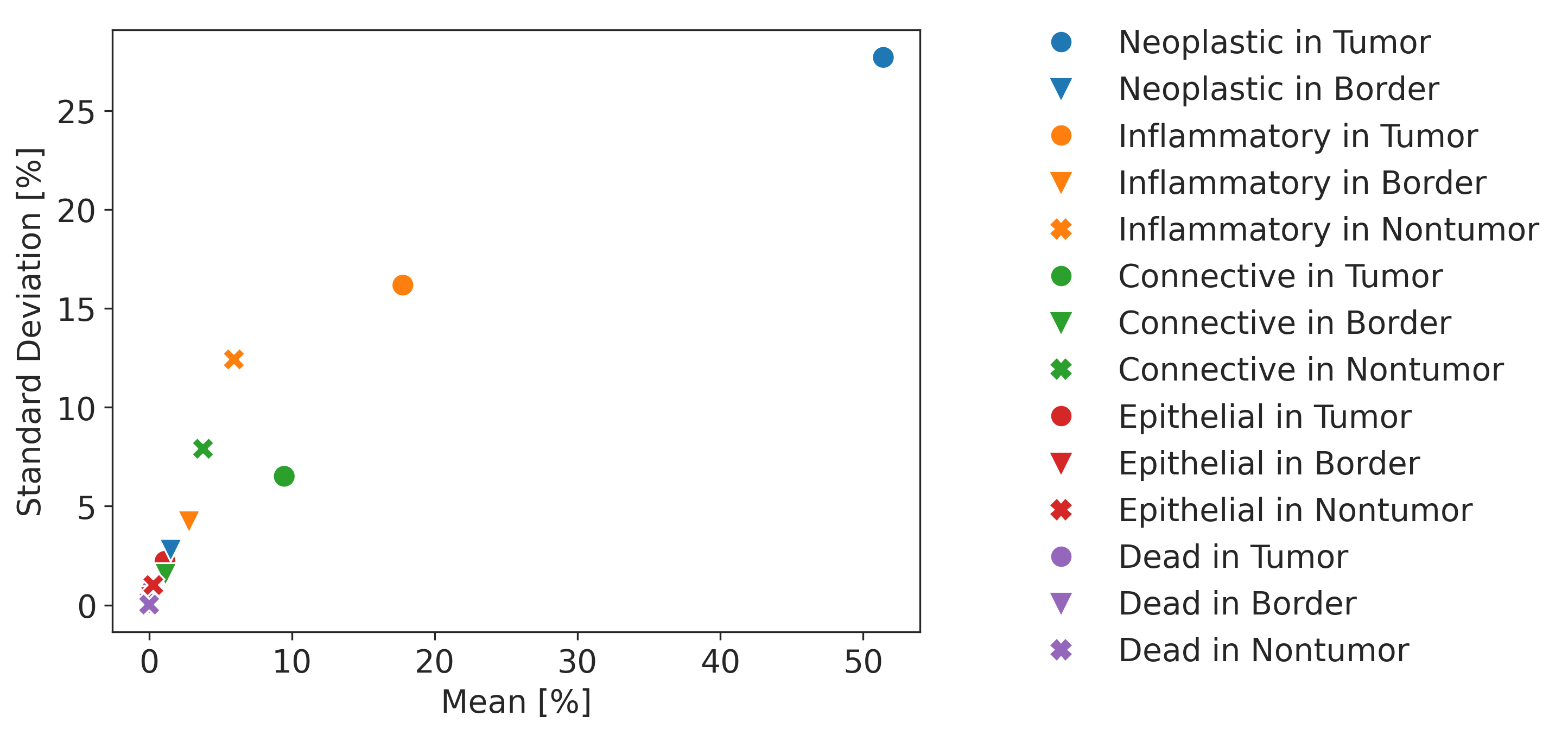}
    \caption{
    \textbf{Most relevant features for HPV prediction in TCGA HNSC}. Mean and standard deviation of the cell type shares per tissue compartment in the top-10\% heatmap regions across all HPV positive samples.}
    \label{fig:feature_relevance}
\end{figure}

\begin{figure}[!t]
    \centering
    \includegraphics[width=0.95\linewidth]{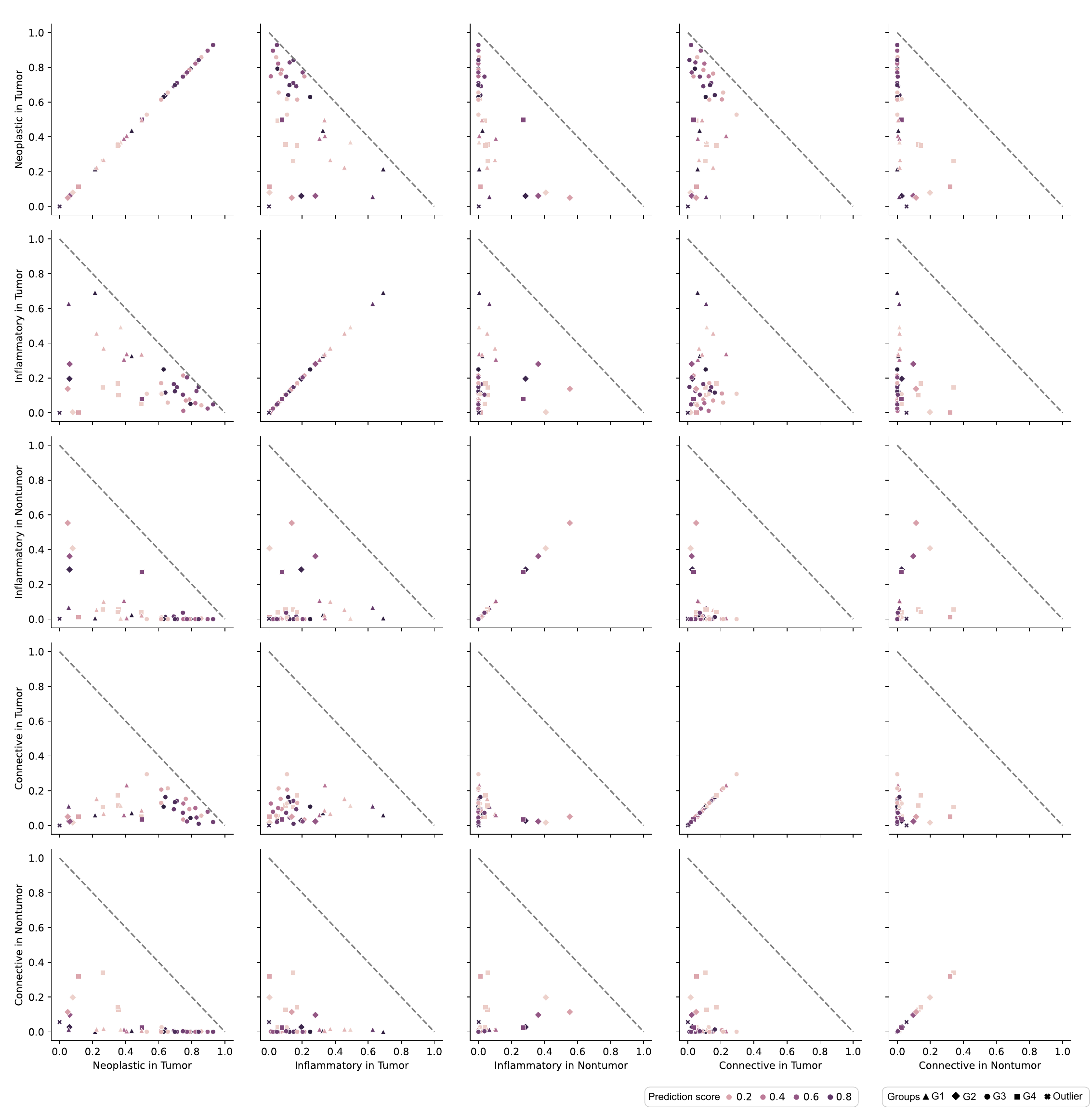}
    \caption{
    \textbf{Comparison of histological feature profiles for HPV prediction in TCGA HNSC}. Scatter matrix displaying the distribution of cell type shares per tissue compartment in the HL10 regions across all HPV positive samples for five selected histological features. The prediction scores indicate the model prediction for these samples (closer to 1 = higher HPV-positive likelihood). The cluster markers represent the clusters revealed with Agglomerative Clustering. See also Figure~\ref{fig:hpv_analysis}-b.}
    \label{supp:fig:scatter_matrix}
\end{figure}


\FloatBarrier

\bibliographystyle{unsrt}

\bibliography{bibliography}

\end{document}